\documentclass[letterpaper]{article} 
\usepackage[preprint]{aaai2027}
\usepackage[hyphens]{url}  
\usepackage{graphicx} 
\usepackage{natbib}  
\usepackage{caption} 
\usepackage{algorithm}
\usepackage{algorithmic}
\usepackage{booktabs}
\usepackage{amsmath}
\usepackage{amssymb}
\usepackage{array}
\usepackage{multirow}
\usepackage{dblfloatfix}
\usepackage{cuted} 

\newcommand{\E}{\mathbb{E}}
\newcommand{\clip}{\operatorname{clip}}

\newcommand{\gbepo}{\textnormal{Gated}-\textsc{BEPO}}

\newcommand{\Succ}{\operatorname{Succ}}

\title{Gated-BEPO: Confidence-Gated Bellman Credit Assignment\\for Large Language Model Agents}
\author{
    Hongxi Yan\textsuperscript{\rm 1,\rm 2},
    Ziyue Huang\textsuperscript{\rm 3},
    Shichao Fan\textsuperscript{\rm 1},
    Qingjie Liu\textsuperscript{\rm 1,\rm 2}
}
\affiliations{
    \textsuperscript{\rm 1}State Key Laboratory of Virtual Reality Technology and Systems,\\
    Beihang University, Beijing, China\\
    \textsuperscript{\rm 2}Zhongguancun Laboratory, Beijing, China\\
    \textsuperscript{\rm 3}Qualcomm, Beijing, China
}

\begin{document}
\raggedbottom

\maketitle
\begingroup
\renewcommand{\thefootnote}{}
\footnotetext{Code: \url{https://github.com/Y-Claw/Gated-BEPO.git}}
\endgroup

\begin{abstract}
Training large language model agents in long-horizon environments requires assigning credit from sparse terminal outcomes to individual actions. Existing critic-free methods propagate trajectory-level rewards uniformly across steps, while recent approaches construct step-level groups by matching repeated states and compare actions within each group. The former cannot distinguish useful actions in failed trajectories from ineffective actions in successful ones. The latter rely on step credit derived directly from individual trajectory outcomes and fixed-weight fusion with episode-level credit. We propose \gbepo{}, which derives step-level credit from empirical rollout graphs. For each rollout group, \gbepo{} constructs an empirical graph and estimates node values through a mean-backup Bellman fixed point that reflects the empirical action distribution of the current policy. We then accumulate these temporal-difference residuals along each sampled trajectory using generalized advantage estimation, yielding step-level Bellman advantages that capture both immediate and downstream effects. To adaptively fuse episode- and step-level credit, a confidence gate incorporates Bellman credit only at states with multiple observed successors and otherwise uses episode-level credit. Experiments on WebShop, ALFWorld, and visual Sokoban show consistent improvements across language and vision-language models, while diagnostic ablations support the effectiveness of Bellman fixed-point value estimation and show that step-level credit should be incorporated selectively rather than uniformly into the final advantage.
\end{abstract}

\section{Introduction}
\begin{figure}[!t]
\centering
\includegraphics[width=\linewidth]{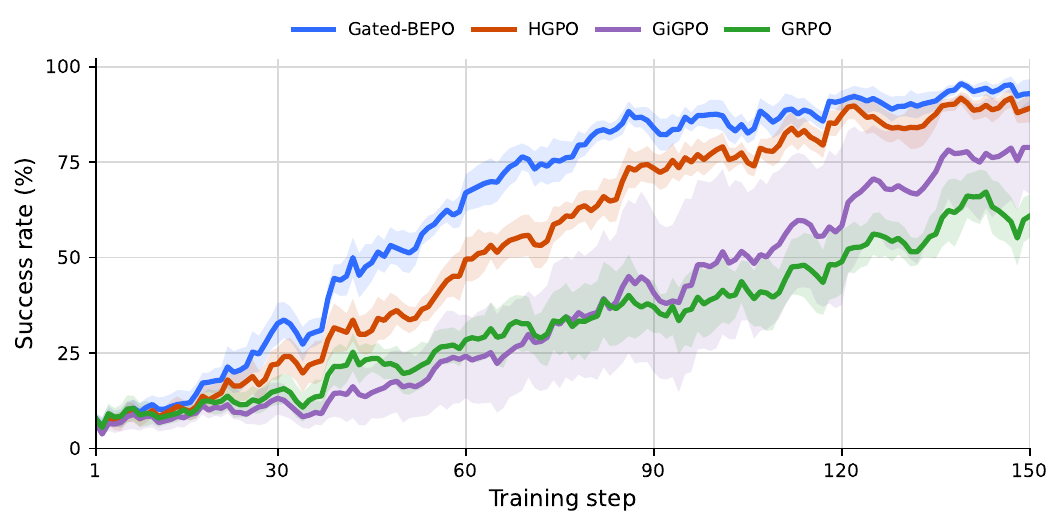}
\caption{ALFWorld success rates (Qwen2.5-1.5B).}
\label{fig:alfworld-success-curves}
\end{figure}

Large language models are increasingly used as interactive agents that navigate web interfaces, manipulate objects in embodied environments, and solve multi-step planning tasks~\citep{yao2023react,qin2023toolllm,shinn2023reflexion,wang2023voyager}. Unlike static text generation, these settings require agents to interact with environments over long horizons before observing sparse task outcomes~\citep{arjona2019rudder,pignatelli2024survey,harutyunyan2019hca}. Credit assignment is therefore central to agent training, since an early action may affect success only after many later interactions~\citep{zhang2026credit,tan2026hindsight}. Critic-free methods such as GRPO avoid the cost and instability of a learned critic, but they usually optimize trajectory-level returns by assigning the same outcome-derived advantage to every step~\citep{shao2024deepseekmath,chen2025loop,yu2025dapo}. This coarse attribution can slow learning and limit final performance because it reinforces ineffective actions in successful trajectories while penalizing useful ones in failed trajectories. As shown in Figure~\ref{fig:alfworld-success-curves}, GRPO improves more slowly and converges to a substantially lower success rate than the other methods.

\begin{figure*}[t]
\centering
\includegraphics[width=0.98\textwidth]{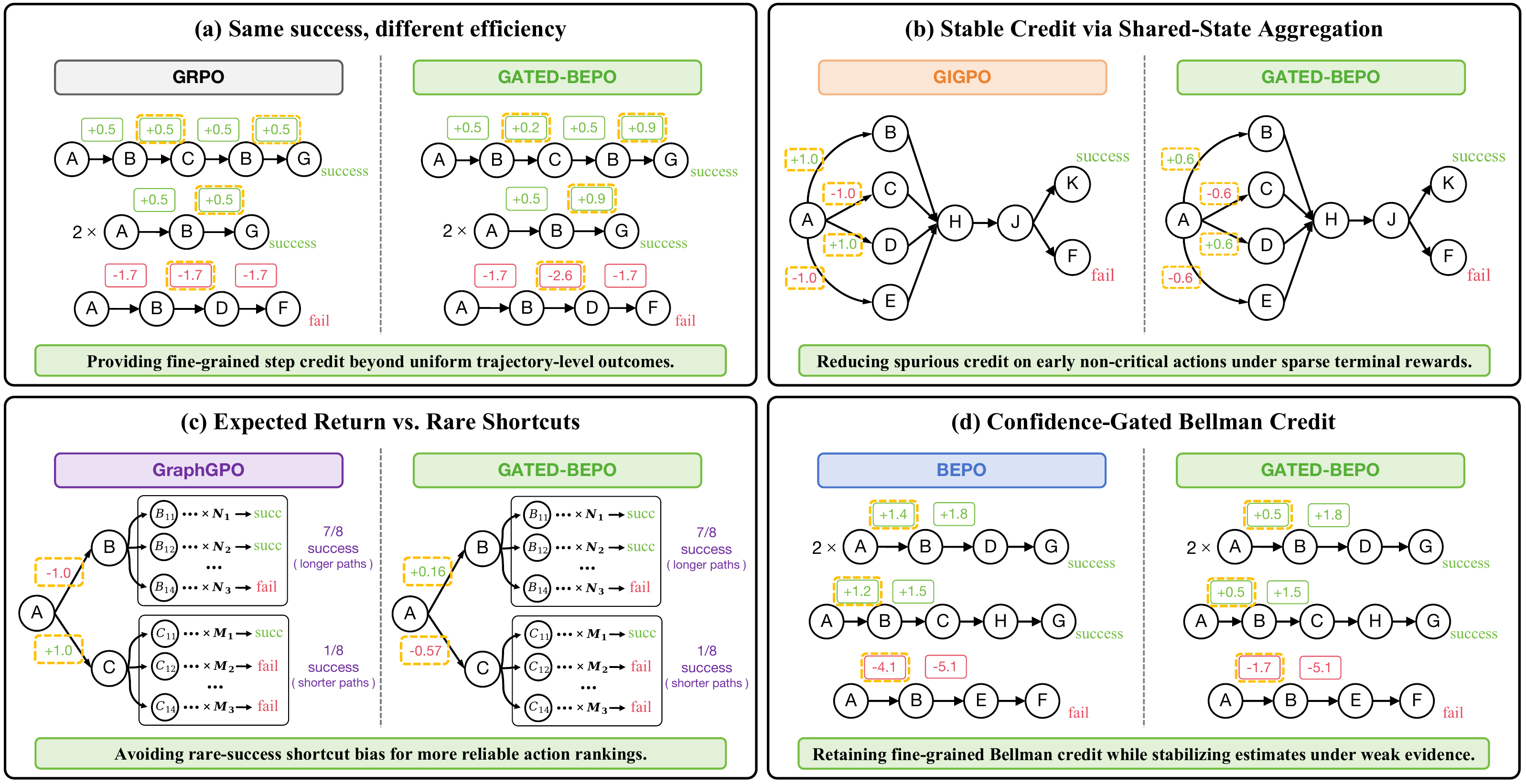}
\caption{Credit assignment across representative critic-free methods. \gbepo{} provides fine-grained step credit, propagates information through shared downstream states, evaluates empirical expected returns, and uses confidence gating to stabilize Bellman credit when local evidence is weak.}
\label{fig:credit-comparison}
\end{figure*}

Therefore, beyond trajectory-level attribution, recent agentic RL methods try to construct fine-grained step-level signals from rollout groups~\citep{zhang2026credit,tan2026hindsight}. Several recent methods exploit repeated observations or histories across rollouts to construct step-level groups for finer-grained credit assignment. GiGPO and HGPO compare actions in the same step group through state matching or history-aware matching~\citep{feng2025gigpo,he2026hgpo}. GAGPO instead constructs a grouped return proxy and propagates it through TD/GAE-style temporal advantages~\citep{zhu2026gagpo}. However, these methods do not explicitly account for the downstream transition structure shared across rollouts. GraphGPO constructs a state-transition graph by merging shared observations across rollouts and uses the reciprocal of the shortest path to an observed successful terminal state as a state-value signal~\citep{cheng2026graphgpo}. However, this construction can suffer from perceptual aliasing when identical observations correspond to different underlying states~\citep{shani2004aliasing}. Moreover, existing methods do not use local comparative evidence to decide when to apply step credit and how much episode credit to retain.

To address these limitations, we propose Gated-BEPO, a gated variant of Bellman-estimated Policy Optimization (BEPO) for critic-free step credit assignment from empirical rollout graphs. For each rollout group, Gated-BEPO merges identical observed states into an empirical state graph and uses a mean-backup Bellman fixed point to estimate values for its graph nodes under the current policy. After Bellman evaluation, Gated-BEPO computes fixed-point temporal-difference residuals and propagates them along each trajectory with generalized advantage estimation. In this way, Gated-BEPO preserves the advantage of graph aggregation by reducing trajectory-level estimation noise and reducing sensitivity to perceptual aliasing.

Furthermore, local value comparisons require empirical alternatives, while their reliability should also determine how strongly they influence policy updates. Gated-BEPO therefore uses the graph-confidence gate to control both the availability and the relative contribution of fixed-point credit. At states without multiple observed successors, the local graph provides no empirical basis for comparing alternative transitions; the fixed-point advantage is therefore suppressed, and learning relies on the full group-relative outcome advantage. At states with observed branching, the fixed-point advantage is activated and assigned greater relative emphasis, while the outcome contribution is reduced but retained as a stabilizing global signal. This confidence-dependent mixture prevents unsupported local estimates from affecting the policy and prevents supported step-level credit from being overwhelmed by trajectory-level outcomes.

To make these distinctions concrete, Table~\ref{tab:method-comparison} summarizes the key differences among representative critic-free step-credit estimators. Figure~\ref{fig:credit-comparison} then contrasts these methods across four rollout patterns, illustrating where uniform outcome credit, state matching, and graph-distance credit can become insufficient. It further shows how \gbepo{} combines behavior-policy evaluation with evidence-gated step credit to address these cases.

\begin{table}[t]
\centering
\footnotesize
\begin{tabular*}{\columnwidth}{@{\extracolsep{\fill}}llcc@{}}
\toprule
Method & Step credit & TD/GAE & Gate/mix \\
\midrule
GiGPO/HGPO & State/history outcomes & No & Fixed \\
GAGPO & Return proxy & Yes & None \\
GraphGPO & Graph distance & No & Fixed \\
\gbepo{} & Mean Bellman & Yes & Adaptive \\
\bottomrule
\end{tabular*}
\caption{Comparison of critic-free step-credit estimators.}
\label{tab:method-comparison}
\end{table}

Overall, our contributions are summarized below.

\begin{figure*}[t]
    \centering
    \includegraphics[width=0.98\textwidth]{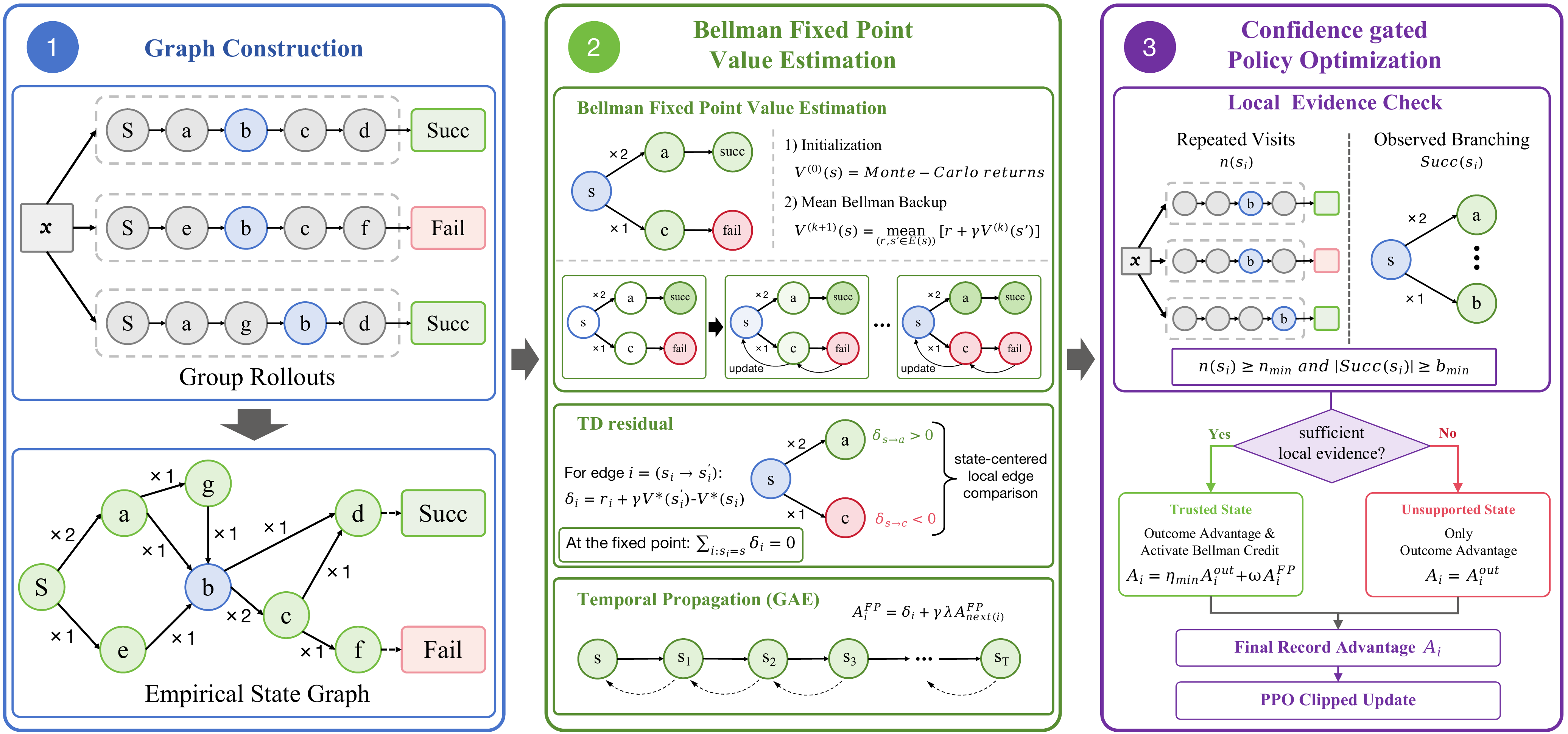}
    \caption{Overview of the \gbepo{} advantage computation pipeline. \gbepo{} builds an empirical rollout graph, derives fixed-point step credit, applies confidence-gated mixing with outcome credit, and broadcasts the resulting record advantage to valid response tokens.}
    \label{fig:method-overview}
\end{figure*}

\begin{itemize}
    \item We propose a critic-free Bellman fixed-point estimator that derives state values and helps reduce noise in credit estimation.
    \item We introduce confidence-gated credit mixing, which uses local graph evidence to gate Bellman credit and determine how to balance outcome- and step-level credit. This prevents unsupported local estimates from introducing noise and reliable step-level credit from being overwhelmed by coarse outcome signals.
    \item We evaluate Gated-BEPO on WebShop, ALFWorld, and visual Sokoban under matched training and evaluation protocols. Experiments show strong performance across Qwen2.5 text and vision-language backbones, with success-rate gains of up to $4.1$\% over the strongest method, and ablations support the importance of the Bellman fixed-point signal, confidence gate, and balanced outcome-step fusion.
\end{itemize}

\section{Related Work}
\paragraph{Outcome-based credit assignment.}
Large language model agents extend language models from static prediction to multi-step interaction with external environments~\citep{liu2023agentbench,schick2023toolformer}. Recent work trains or self-improves web agents using self-generated interaction trajectories and online reinforcement learning~\citep{patel2024selfimprove,qi2025webrl,wei2025webagentr1}. Such training often relies on trajectory-level outcomes~\citep{christiano2017preferences,stiennon2020summarize,ouyang2022instructgpt}. Critic-free group-relative objectives avoid a learned critic and reduce training cost, but they assign credit according to final returns~\citep{ahmadian2024back}. Process supervision can provide finer step-level labels~\citep{lightman2023verify}, but it requires external annotations or a process reward model. These limitations motivate a critic-free approach that can recover informative step-level credit from sparse environment rewards and the agent's own rollout structure, without human step labels or auxiliary learned reward or value models.

\paragraph{Step-level credit assignment.}
Recent agentic RL methods seek finer credit assignment while preserving the advantages of group-based optimization~\citep{kazemnejad2024vineppo}. Turn-level reward design and hierarchical trajectory decomposition derive finer-grained signals for multi-turn or structurally complex agents~\citep{wei2025turnreward,luo2025agentlightning}. GiGPO~\citep{feng2025gigpo} constructs step-level groups by matching repeated anchor states and combines episode-level and step-level relative advantages. HGPO~\citep{he2026hgpo} addresses context inconsistency by assigning steps to multiple history-aware groups and adaptively aggregating their advantages. GAGPO~\citep{zhu2026gagpo} builds a grouped return proxy and uses it in TD/GAE-style temporal advantages. GraphGPO~\citep{cheng2026graphgpo} aggregates rollouts into a state-transition graph and assigns edge credit according to progress toward the goal under a graph-distance objective. Collectively, these methods either derive step credit from trajectory returns without modeling shared downstream transition structure, or rely on optimistic graph-distance values that can be brittle under perceptual aliasing. They also lack an evidence-based mechanism for jointly deciding when local step credit is reliable and how much episode-level credit should be retained.

\section{Method}

\subsection{Problem Setup}
We consider a large language model agent parameterized by $\pi_\theta$. For an initial task instance $x$, the agent generates an environment action at each interaction step and receives the next textual observation and an environment reward. A rollout trajectory is
\begin{equation}
\tau=(s_0,a_0,r_0,s_1,\ldots,s_{T-1},a_{T-1},r_{T-1},s_T),
\end{equation}
where $s_t$ is a hashable key derived from the environment-provided observation (text in text-based environments and an image in visual environments), $a_t$ is one complete language-model response interpreted as an environment action, and $r_t$ is the transition reward. For every task instance, we sample a rollout group of $K$ trajectories from the same initial state. We refer to each environment step $(s_t,a_t,r_t,s_{t+1})$ as a rollout record. Its scalar advantage is shared by the valid response tokens that form $a_t$.

\subsection{Method Overview}
As illustrated in Figure~\ref{fig:method-overview}, \gbepo{} combines a group-relative outcome advantage with a fixed-point step advantage through confidence-gated mixing. The gate leaves the method to rely entirely on outcome credit when no distinct successors are observed. Once observed local comparative evidence, it activates the fixed-point advantage and down-weights the outcome contribution.

For a rollout record $i$, let $u(i)$ identify its task-level rollout group, $s_i$ its source state, and $s'_i$ its successor (or an absorbing terminal state $\bot$). We define its mixed advantage as
\begin{equation}
\widehat A_i = \eta_i \widehat A_i^{\mathrm{out}}
+ w\,\rho(s_i)\widehat A_i^{\mathrm{FP}},
\label{eq:combined_advantage}
\end{equation}
where $\widehat A_i^{\mathrm{out}}$ is a group-relative outcome advantage, $\widehat A_i^{\mathrm{FP}}$ is a fixed-point step advantage, $\rho(s_i)\in\{0,1\}$ is a graph-confidence gate, $\eta_i$ is a complementary outcome weight, and $w$ controls the contribution of the step signal.

\subsection{Empirical Graph and Bellman Fixed Point}
For each rollout group independently, we merge identical observations into a shared state. The resulting empirical graph retains the observed transition multiplicities. Specifically, let
\begin{equation}
E(s)=\{(r_i,s'_i):s_i=s\}
\end{equation}
be the multiset of outgoing transitions from state $s$. Multiple records with the same start and end states are counted as distinct edges.

We estimate state values directly from the empirical graph, without a learned critic. Terminal transitions lead to outcome-typed absorbing states representing success, failure, or truncation, whose values are set to zero. We update each nonterminal state using the following mean Bellman backup~\citep{sutton1988td}:
\begin{equation}
V^{(k+1)}(s)=\frac{1}{|E(s)|}\sum_{(r,s')\in E(s)}\left[r+\gamma V^{(k)}(s')\right].
\label{eq:bellman_backup}
\end{equation}
where $k$ denotes the Bellman-backup iteration. Averaging these Bellman targets over the transition multiset $E(s)$ estimates the expected return under the current policy. With $\gamma<1$, the operator in Equation~\ref{eq:bellman_backup} is a contraction on the finite empirical graph and therefore has a unique fixed point. We initialize the iteration with Monte-Carlo returns and run the Bellman update until convergence or until the iteration budget is reached.

For an rollout record $i$, the fixed-point temporal-difference residual is
\begin{equation}
\delta_i=r_i+\gamma V(s'_i)-V(s_i).
\label{eq:td_residual}
\end{equation}
At the fixed point, residuals are centered within every state in the empirical graph as follows.
\begin{equation}
\sum_{i:s_i=s}\delta_i=0.
\label{eq:state_centering}
\end{equation}
Thus, before temporal propagation, the residual measures whether an observed edge is better or worse than the empirical average of outgoing edges from the same state, providing a critic-free, state-centered local advantage signal for policy optimization~\citep{williams1992reinforce,sutton2000policygradient}.

We propagate residuals along each original trajectory using GAE~\citep{schulman2016gae}.
\begin{equation}
\widehat A_i^{\mathrm{FP}}=
\delta_i+\gamma\lambda\widehat A_{\operatorname{next}(i)}^{\mathrm{FP}},
\label{eq:fp_gae}
\end{equation}
where the continuation is zero after termination. We compute this recursion before applying the gate. Consequently, $\rho(s_i)$ determines whether the accumulated fixed-point advantage is used at record $i$, but it does not remove $\delta_i$ from the GAE targets of earlier records. This preserves delayed credit from later transitions even when their own gates are closed. We then standardize $\widehat A^{\mathrm{FP}}$ over all records in each rollout group before gating and mixing, so the normalization statistics include both open- and closed-gate records.

\subsection{Confidence-Gated Credit Mixing}
A local graph signal is only useful when the rollout group supplies empirical alternatives. We therefore define the visit count and the number of distinct observed successors as
\begin{equation}
n(s)=|\{i:s_i=s\}|,\qquad
\Succ(s)=\{s'_i:s_i=s\}.
\end{equation}
We define the binary graph-confidence gate as
\begin{equation}
\rho(s)=
\begin{cases}
1, & n(s)\geq n_{\min}\ \text{and}\ |\Succ(s)|\geq b_{\min},\\
0, & \text{otherwise}.
\end{cases}
\label{eq:confidence_gate}
\end{equation}
Because every distinct successor requires an observed outgoing transition, $|\Succ(s)|\leq n(s)$. Therefore, with the default $n_{\min}=b_{\min}=2$, Equation~\ref{eq:confidence_gate} reduces to requiring at least two distinct successors, which both implies repeated visits and supplies empirical local alternatives. We retain $n_{\min}$ as a separate threshold to test whether additional visits beyond this minimum improve coverage. Outcome-typed terminal states count as distinct successors.

The outcome advantage receives full weight when the gate is closed and is down-weighted when the gate is open:
\begin{equation}
\eta_i=\eta_{\min}+(1-\eta_{\min})(1-\rho(s_i)).
\label{eq:outcome_weight}
\end{equation}
Reducing outcome credit prevents it from dominating the mixed advantage and helps keep advantage scales comparable between gated and ungated states.

\begin{table*}[!t]
\centering
\small
\setlength{\tabcolsep}{1pt}
\begin{tabular}{llccccccccc}
\toprule
\multirow{2}{*}{Type} & \multirow{2}{*}{Method}
& \multicolumn{7}{c}{ALFWorld}
& \multicolumn{2}{c}{WebShop} \\
\cmidrule(lr){3-9}\cmidrule(lr){10-11}
& & Pick & Look & Clean & Heat & Cool & Pick2 & All & Score & Succ. \\
\midrule
\multicolumn{11}{l}{\textit{Closed-source model}} \\
Prompting & GPT-4o & $75.3$ & $60.8$ & $31.2$ & $56.7$ & $21.6$ & $49.8$ & $48.0$ & $31.8$ & $23.7$ \\
Prompting & Gemini-2.5-Pro & $92.8$ & $63.3$ & $62.1$ & $69.0$ & $26.6$ & $58.7$ & $60.3$ & $42.5$ & $35.9$ \\
\midrule
\multicolumn{11}{l}{\textit{Qwen2.5-1.5B-Instruct}} \\
Prompting & Qwen2.5 & $5.9$ & $5.5$ & $3.3$ & $9.7$ & $4.2$ & $0.0$ & $4.1$ & $23.1$ & $5.2$ \\
Prompting & ReAct & $17.4$ & $20.5$ & $15.7$ & $6.2$ & $7.7$ & $2.0$ & $12.8$ & $40.1$ & $11.3$ \\
Prompting & Reflexion & $35.3$ & $22.2$ & $21.7$ & $13.6$ & $19.4$ & $3.7$ & $21.8$ & $55.8$ & $21.9$ \\
RL Training & PPO (with critic) & $64.8{\pm}3.5$ & $40.5{\pm}6.9$ & $57.1{\pm}4.9$ & $60.6{\pm}6.6$ & $46.4{\pm}4.0$ & $47.4{\pm}1.9$ & $54.4{\pm}3.1$ & $73.8{\pm}3.0$ & $51.5{\pm}2.9$ \\
RL Training & RLOO & $88.3{\pm}3.0$ & $52.8{\pm}8.6$ & $71.0{\pm}5.9$ & $62.8{\pm}8.7$ & $66.4{\pm}5.5$ & $56.9{\pm}4.7$ & $69.7{\pm}2.5$ & $73.9{\pm}5.6$ & $52.1{\pm}6.7$ \\
RL Training & GRPO & $85.3{\pm}1.5$ & $53.7{\pm}8.0$ & $84.5{\pm}6.8$ & $78.2{\pm}7.9$ & $59.7{\pm}5.0$ & $53.5{\pm}5.6$ & $72.8{\pm}3.6$ & $75.8{\pm}3.5$ & $56.8{\pm}3.8$ \\
RL Training & GAGPO & $91.9{\pm}1.4$ & $51.9{\pm}10.5$ & $91.7{\pm}2.7$ & $93.2{\pm}5.3$ & $88.3{\pm}3.1$ & $76.3{\pm}5.0$ & $85.5{\pm}0.9$ & $80.8{\pm}5.1$ & $59.5{\pm}3.6$ \\
RL Training & GiGPO & $94.4{\pm}5.9$ & $67.5{\pm}4.6$ & $94.8{\pm}3.8$ & $94.4{\pm}7.8$ & $79.8{\pm}4.7$ & $76.4{\pm}5.4$ & $86.7{\pm}1.7$ & $83.1{\pm}1.6$ & $65.0{\pm}3.2$ \\
RL Training & HGPO & $94.3{\pm}2.9$ & $69.1{\pm}9.2$ & $95.9{\pm}2.8$ & $97.4{\pm}3.6$ & $92.0{\pm}2.3$ & $82.3{\pm}3.2$ & $90.5{\pm}0.4$ & $85.5{\pm}0.5$ & $70.5{\pm}1.7$ \\
RL Training & \gbepo{} & $\mathbf{96.6{\pm}2.5}$ & $\mathbf{81.5{\pm}10.5}$ & $\mathbf{98.9{\pm}0.9}$ & $\mathbf{98.3{\pm}1.2}$ & $\mathbf{92.6{\pm}2.6}$ & $\mathbf{84.4{\pm}5.9}$ & $\mathbf{93.2{\pm}2.2}$ & $\mathbf{87.8{\pm}0.6}$ & $\mathbf{74.0{\pm}1.4}$ \\
\midrule
\multicolumn{11}{l}{\textit{Qwen2.5-7B-Instruct}} \\
Prompting & Qwen2.5 & $33.4$ & $21.6$ & $19.3$ & $6.9$ & $2.8$ & $3.2$ & $14.8$ & $26.4$ & $7.8$ \\
Prompting & ReAct & $48.5$ & $35.4$ & $34.3$ & $13.2$ & $18.2$ & $17.6$ & $31.2$ & $46.2$ & $19.5$ \\
Prompting & Reflexion & $62.0$ & $41.6$ & $44.9$ & $30.9$ & $36.3$ & $23.8$ & $42.7$ & $58.1$ & $28.8$ \\
RL Training & PPO (with critic) & $92.3{\pm}4.0$ & $64.0{\pm}8.4$ & $92.5{\pm}2.4$ & $89.5{\pm}7.0$ & $80.3{\pm}2.0$ & $68.8{\pm}8.3$ & $80.4{\pm}2.7$ & $81.4{\pm}3.1$ & $68.7{\pm}5.1$ \\
RL Training & RLOO & $87.6{\pm}4.3$ & $78.2{\pm}8.3$ & $87.3{\pm}5.8$ & $81.3{\pm}7.6$ & $71.9{\pm}5.2$ & $48.9{\pm}8.4$ & $75.5{\pm}4.6$ & $80.3{\pm}3.2$ & $65.7{\pm}4.0$ \\
RL Training & GRPO & $90.8{\pm}5.1$ & $66.1{\pm}6.7$ & $89.3{\pm}5.4$ & $74.7{\pm}6.9$ & $72.5{\pm}5.4$ & $64.7{\pm}7.3$ & $77.6{\pm}5.2$ & $79.3{\pm}2.8$ & $66.1{\pm}3.7$ \\
RL Training & GAGPO & $92.9{\pm}10.0$ & $71.6{\pm}17.2$ & $92.4{\pm}10.7$ & $82.9{\pm}12.3$ & $91.4{\pm}4.4$ & $84.0{\pm}6.4$ & $88.3{\pm}8.7$ & $84.8{\pm}2.7$ & $66.5{\pm}6.0$ \\
RL Training & GiGPO & $97.7{\pm}1.6$ & $82.7{\pm}7.9$ & $\mathbf{98.8{\pm}1.6}$ & $83.7{\pm}7.2$ & $89.3{\pm}8.2$ & $79.2{\pm}6.6$ & $90.8{\pm}1.3$ & $84.4{\pm}2.9$ & $72.8{\pm}3.2$ \\
RL Training & HGPO & $99.7{\pm}0.5$ & $\mathbf{84.0{\pm}1.8}$ & $95.0{\pm}1.4$ & $94.0{\pm}2.4$ & $82.1{\pm}9.1$ & $86.1{\pm}5.3$ & $91.8{\pm}1.8$ & $86.0{\pm}0.7$ & $72.8{\pm}1.6$ \\
RL Training & \gbepo{} & $\mathbf{99.7{\pm}0.5}$ & $71.6{\pm}1.7$ & $97.7{\pm}0.9$ & $\mathbf{95.7{\pm}3.2}$ & $\mathbf{92.6{\pm}2.6}$ & $\mathbf{93.9{\pm}3.2}$ & $\mathbf{94.7{\pm}0.7}$ & $\mathbf{87.2{\pm}0.3}$ & $\mathbf{77.0{\pm}0.8}$ \\
\bottomrule
\end{tabular}
\caption{Main results on ALFWorld and WebShop. For ALFWorld, we report each subtask and the overall average. For WebShop, we report task score and success rate.}
\label{tab:main-results}
\end{table*}

\subsection{Outcome Advantage and Optimization}
Let $R(\tau_i)$ denote the total environment return of the trajectory containing record $i$. We compute the group-relative outcome advantage as
\begin{equation}
\widehat A_i^{\mathrm{out}}=
\frac{R(\tau_i)-\mu_{u(i)}^R}{\sigma_{u(i)}^R+\epsilon},
\label{eq:outcome_advantage}
\end{equation}
where $\mu_{u(i)}^R$ and $\sigma_{u(i)}^R$ are computed over rollout records in the same task-level group after assigning each trajectory return to all of its records. The resulting normalization is therefore record-weighted rather than trajectory-weighted. After mixing, the record advantage is broadcast to its valid response tokens. \gbepo{} changes only the advantage estimator. Let $q_{i,\ell}(\theta)$ denote the standard PPO importance ratio for response token $y_{i,\ell}$. The actor uses the clipped objective~\citep{schulman2017ppo}
\begin{equation}
\begin{aligned}
\mathcal{L}_{\mathrm{clip}}(\theta)
&=\E\Bigl[\min\bigl(
q_{i,\ell}\widehat A_i,\\
&\qquad
\clip(q_{i,\ell},1-\varepsilon,1+\varepsilon)\widehat A_i
\bigr)\Bigr].
\end{aligned}
\end{equation}

\section{Experiments}

\subsection{Experimental Setup}
We evaluate \gbepo{} on WebShop~\citep{yao2022webshop}, ALFWorld~\citep{shridhar2021alfworld}, and visual Sokoban~\citep{junghanns2001sokoban}. WebShop and ALFWorld use Qwen2.5-1.5B-Instruct and Qwen2.5-7B-Instruct~\citep{yang2025qwen25}, while Sokoban uses Qwen2.5-VL-3B. WebShop and ALFWorld use rollout groups of $K=8$ and train for $150$ optimization steps. Episodes are limited to $15$ steps in WebShop and $50$ steps in ALFWorld. All main results use three independent training seeds, each evaluated and averaged over three environment seeds. Full hyperparameters are provided in the supplementary material.

\subsection{Main Results}
Table~\ref{tab:main-results} summarizes the main results on ALFWorld and WebShop. 
Against HGPO, \gbepo{} improves WebShop success by $3.51\%$ and $4.17\%$ and ALFWorld overall success by $2.78\%$ and $2.95\%$ at 1.5B and 7B, respectively, while achieving the best WebShop task scores. It also exceeds GAGPO on ALFWorld by $7.7\%$ and $6.4\%$ in success at 1.5B and 7B, respectively. On WebShop, it exceeds GAGPO by $14.5\%$ and $10.5\%$ in success and by $6.9\%$ and $2.4\%$ in task score at 1.5B and 7B, respectively. The same ordering transfers to Qwen3-1.7B (provided in supplementary material), where \gbepo{} outperforms HGPO by $7.29\%$ and $4.86\%$ in success on ALFWorld and WebShop, respectively.

Figure~\ref{fig:alfworld-success-curves} shows that, on ALFWorld at 1.5B, \gbepo{} improves training success faster than HGPO, GiGPO, and GRPO during the early and middle stages of training. It also converges to the highest training success rate. This pattern suggests that gated Bellman credit provides a more informative step-level signal from sparse trajectory outcomes. The 7B curves in the supplementary material show a similar advantage in learning speed, with \gbepo{} reaching the same success rates earlier than HGPO. In terms of final performance, \gbepo{} converges to a higher validation success rate than HGPO.
 
\begin{table}[t]
\centering
\small
\setlength{\tabcolsep}{5pt}
\begin{tabular}{llc}
\toprule
Type & Method & Sokoban $6{\times}6$ Succ. \\
\midrule
Prompting & Qwen2.5-VL & $11.70$ \\
RL Training & GRPO & $67.1{\pm}4.7$ \\
RL Training & GiGPO & $76.9{\pm}2.7$ \\
RL Training & \gbepo{} & $\mathbf{80.0{\pm}3.2}$ \\
\bottomrule
\end{tabular}
\caption{Sokoban success rate (\%) on $6{\times}6$ visual tasks.}
\label{tab:sokoban-main}
\end{table}

Table~\ref{tab:sokoban-main} reports the visual Sokoban result separately. All three RL methods achieve strong performance, with \gbepo{} reaching $80.03\pm0.32\%$ success and exceeding the reported GRPO and GiGPO means by $12.93\%$ and $3.13\%$, respectively. These results suggest that the gated Bellman signal transfers effectively to a vision-language planning setting.

Additionally, \gbepo{} adds little computational overhead. On ALFWorld, the complete advantage stage takes $0.361$ seconds per update, while the remaining rollout and optimization stages take about $225.7$ seconds in total (full time cost provided in the supplementary material).

\subsection{Observation Aliasing}
\begin{figure}[!t]
\centering
\includegraphics[width=\linewidth]{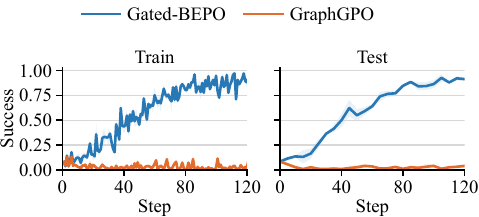}
\caption{ALFWorld training and test success rates under the matched raw-observation protocol.}
\label{fig:alfworld-aliasing-curves}
\end{figure}

We evaluate GraphGPO under two ALFWorld protocols. The matched protocol uses the raw observations used by the other baselines, while the original GraphGPO protocol augments observations with environment-specific state information, such as location, inventory, object history, and admissible actions (details are provided in the supplementary material). With this augmented representation, GraphGPO reaches $92.7\%$ success, below \gbepo{}'s $95.7\%$. Under the matched raw-observation protocol in Figure~\ref{fig:alfworld-aliasing-curves}, GraphGPO nearly fails to learn while \gbepo{} remains effective. This failure arises because GraphGPO estimates credit solely from shortest paths on the merged observation graph and therefore cannot distinguish identical observations that correspond to different hidden states. Although \gbepo{} also merges identical observations, its TD residuals are computed on observed transitions and propagated along the original trajectories with GAE, so occurrences of the same observation can still receive different credit when their successors and returns differ.

\subsection{Ablation Studies}
Given the substantial cost of extensive ablations and the common use of a fixed seed results in prior work, unless otherwise stated, our ablation experiments use Qwen2.5-1.5B with a fixed training seed.

\begin{table}[!t]
\centering
\small
\setlength{\tabcolsep}{2.2pt}
\begin{tabular}{lccccc}
\toprule
Setting & FP & Gate & $\downarrow$Out. & Success & Score \\
\midrule
Outcome only & $\times$ & -- & $\times$ & $62.50\pm1.99$ & $81.38\pm0.48$ \\
$+$ FP credit & $\checkmark$ & $\times$ & $\times$ & $66.41\pm3.38$ & $87.16\pm0.41$ \\
$+$ Gate & $\checkmark$ & $\checkmark$ & $\times$ & $73.83\pm2.30$ & $87.26\pm1.01$ \\
\gbepo{} & $\checkmark$ & $\checkmark$ & $\checkmark$ & $\mathbf{75.91\pm2.71}$ & $\mathbf{88.32\pm0.89}$ \\
\bottomrule
\end{tabular}
\caption{Component-wise ablation on WebShop (\%). FP denotes fixed-point credit and $\downarrow$Out.\ denotes trusted-state outcome down-weighting.}
\label{tab:fusion-ablation}
\end{table}

\subsubsection{RQ1. Which components drive Gated-BEPO's gains?}
\paragraph{Component-wise decomposition.}
Table~\ref{tab:fusion-ablation} presents a cumulative ablation of the three core components under a shared computational framework. Starting from outcome-only optimization, the subsequent rows cumulatively add fixed-point credit, confidence gating, and trusted-state outcome down-weighting. Additional outcome-weighting variants are reported in the supplementary material.

Starting from outcome-only optimization, adding the fixed-point step advantage improves success from $62.5\%$ to $66.4\%$. Restricting the same signal to branch-supported states further improves success to $73.8\%$, suggesting a benefit from confidence gating. Finally, down-weighting outcome credit at trusted states yields $75.9\%$, showing an additional benefit from adaptive credit mixing. Overall, these consistent gains demonstrate that all three components play important roles in the final performance.

\paragraph{General effectiveness of confidence gating.}
To test whether confidence-gated mixing is specific to our fixed-point estimator, we apply the same binary gate and dynamic mixing rule to GAGPO while retaining its original return-based step estimator. In WebShop, gating improves GAGPO success from $64.5\%$ to $70.1\%$ and task score from $85.6$ to $86.5$. This controlled comparison provides evidence that selective step-credit integration can be beneficial beyond Bellman fixed-point credit.

\subsubsection{RQ2. What makes a reliable confidence gate?}
\begin{figure}[t]
\centering
\includegraphics[width=0.92\columnwidth]{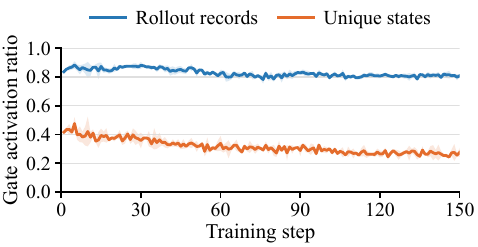}
\caption{WebShop gate coverage over three seeds, measured over rollout records and unique empirical-graph states.}
\label{fig:gate-activation}
\end{figure}

\paragraph{Shared-state coverage.}
Figure~\ref{fig:gate-activation} shows that shared, branching states are common in WebShop. At the final step, the gate covers $80.8\%$ of rollout records but only $27.5\%$ of unique states. Thus, recurrent states supply a large fraction of updates while the gate remains selective. ALFWorld shows the same pattern ($72.0\%$ versus $49.8\%$); complete diagnostics are provided in the supplementary material. This substantial coverage provides the empirical basis for using shared states to construct local comparisons.

\paragraph{Local comparison evidence.}

\begin{table}[!t]
\centering
\small
\setlength{\tabcolsep}{3pt}
\begin{tabular}{lcccc}
\toprule
Gate & $n_{\min}$ & $b_{\min}$ & Success & Score \\
\midrule
All states & $1$ & $1$ & $70.44\pm1.12$ & $85.87\pm0.81$ \\
Repeat only & $2$ & $1$ & $62.50\pm1.91$ & $83.09\pm0.86$ \\
Default & $2$ & $2$ & $\mathbf{75.91\pm2.71}$ & $\mathbf{88.32\pm0.89}$ \\
Strict visits & $3$ & $2$ & $72.01\pm1.57$ & $87.21\pm0.29$ \\
Strict branch & $2$ & $3$ & $66.80\pm2.09$ & $84.25\pm0.90$ \\
\bottomrule
\end{tabular}
\caption{Gate thresholds on WebShop (\%). $b_{\min}$ counts distinct successors.}
\label{tab:gate-ablation}
\end{table}

Table~\ref{tab:gate-ablation} varies the local evidence required by the binary gate. Repeat-only gating obtains only $62.5\%$ success, below the $70.4\%$ achieved by activating the gate at all states, showing that repeated visits without alternative successors do not support a meaningful local comparison. The default gate instead requires two distinct successors and achieves the best success of $75.9\%$. Raising the visit requirement lowers success to $72.0\%$, while requiring three distinct successors reduces it further to $66.8\%$. Continuous alternatives that gradually increase confidence with visit or branching evidence likewise underperform the binary gate (complete definitions and results are in the supplementary material). Together with the broad shared-state coverage above, these results identify observed branching as a useful minimum criterion that filters unsupported comparisons without unnecessarily suppressing usable step credit.

\begin{figure}[!t]
\centering
\includegraphics[width=\linewidth]{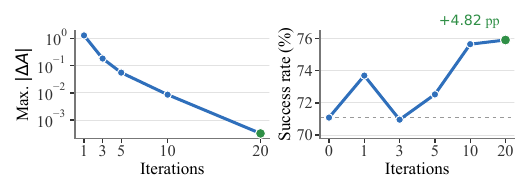}
\caption{Bellman-iteration diagnostics on WebShop. Left: maximum absolute change in the resulting advantage. Right: success rate under each iteration budget.}
\label{fig:ablation-diagnostics}
\end{figure}

\subsubsection{RQ3. How should the Bellman fixed point be computed?}
\paragraph{Backup operator.}
The supplementary material compares backup operators on ALFWorld. Mean backup evaluates the empirical transition distribution, whereas max and softmax emphasize high-value observed successors. Mean achieves $90.4\pm1.0\%$ success  compared with $65.4\pm1.3\%$ for max and $64.6\pm1.0\%$ for softmax. These substantial gaps suggest that planning-style backups amplify rare or aliased transitions and make state-centered comparison less reliable.

\paragraph{Iteration budget.}
Figure~\ref{fig:ablation-diagnostics} examines how the Bellman iteration budget affects the resulting step signal. One iteration improves success from $71.1\%$ to $73.7\%$, indicating that Bellman propagation is already useful, yet additional iterations before reaching a stable fixed point can introduce noise into the TD residuals and step advantages, weakening the gain and yielding $71.0\%$ success at three iterations. As the values stabilize, success rises to $75.9\%$ at 20 iterations. 

\subsubsection{RQ4. How should the step signal be propagated and weighted?}
\begin{table}[!t]
\centering
\small
\setlength{\tabcolsep}{3pt}
\begin{tabular}{lcc}
\toprule
Recursion & Success & Score \\
\midrule
Stop at unsupported & $69.92\pm2.23$ & $87.13\pm0.29$ \\
Residual masking & $72.79\pm2.24$ & $87.49\pm1.21$ \\
Post-GAE gate (ours) & $\mathbf{75.26\pm3.19}$ & $\mathbf{89.92\pm1.00}$ \\
\bottomrule
\end{tabular}
\caption{Gate--GAE propagation ablation on WebShop (\%).}
\label{tab:gate-gae-ablation}
\end{table}

\paragraph{Downstream credit propagation.}
Table~\ref{tab:gate-gae-ablation} tests whether residuals from unsupported downstream states should enter an earlier trusted state's GAE target. Our post-GAE gate performs best. Masking each unsupported residual before GAE lowers success from $75.3$ to $72.8$, while stopping propagation whenever the next state is unsupported further lowers it to $69.9$. Thus, although the gate certifies the location where fixed-point credit is used rather than every residual in its long-range target, retaining downstream residuals provides useful delayed credit. 

\paragraph{Propagation horizon.}
The supplementary GAE-$\lambda$ ablation varies the decay used after computing fixed-point residuals. The two extremes perform poorly: one-step TD credit ($\lambda=0.0$) and full-trajectory propagation ($\lambda=1.0$) achieve $64.6\%$ and $68.9\%$ success, respectively. The intermediate settings perform better, with $\lambda=0.5$ reaching $72.1\%$ and the default $\lambda=0.8$ achieving the best result of $75.9\%$. These results show that a moderate propagation horizon best balances delayed credit assignment against noise accumulation.


\paragraph{Step-credit weighting.}
The supplementary step-weight ablation shows that success increases from $72.0\%$ at $w=0.5$ to $74.0\%$ at $w=1.0$ and peaks at $75.9\%$ with $w=1.5$, before dropping to $71.2\%$ at $w=2.0$. Because the step advantage is standardized before mixing, these results show that a moderate relative weight allows local Bellman comparisons to meaningfully influence the update without dominating the retained outcome credit.

\FloatBarrier

\section{Conclusion}
We introduced \gbepo{} to address the coarse credit produced by sparse trajectory outcomes without training a critic. Its Bellman fixed-point estimator provides more accurate step-level credit and reduces sensitivity to such aliasing. Confidence-gated mixing further prevents weakly supported local estimates from corrupting policy updates while keeping step advantages from being overwhelmed by episode-level advantages. Experiments on WebShop, ALFWorld, and visual Sokoban demonstrate strong performance across language and vision-language agents, and diagnostic ablations support the contribution of each component.

\bibliography{aaai2027}

\clearpage
\appendix
\setcounter{secnumdepth}{1}

\section{Reproducibility Details}

\subsection{Environment Details}

\paragraph{WebShop.}
WebShop is a text-based web-shopping benchmark in which an agent must find and purchase a product that satisfies a natural-language instruction specifying the desired item and attributes~\citep{yao2022webshop}. Each episode starts from the search page. The agent observes the instruction, the current page rendered as text, and the currently admissible interactions. Its action is either \texttt{search[keywords]}, which submits a query, or \texttt{click[value]}, which follows a link, selects a product option, navigates between pages, or chooses \texttt{Buy Now}. Selecting \texttt{Buy Now} terminates the episode; otherwise, we truncate it after 15 environment steps. The original benchmark returns a graded match score $q\in[0,1]$ based on the purchased product and requested attributes, which we retain as TaskScore for evaluation. For policy optimization, the terminal reward is $10$ only when $q=1$ and is zero otherwise, so the binary success event requires complete satisfaction of the shopping instruction.

\paragraph{ALFWorld.}
ALFWorld aligns text-based household tasks with the embodied ALFRED task family, enabling agents to interact through textual observations and high-level commands~\citep{shridhar2021alfworld}. We use its TextWorld interface with raw textual observations and the in-distribution evaluation split. Episodes cover six task types: placing one object in a receptacle, placing two objects in a receptacle, examining an object under a light, and heating, cooling, or cleaning an object before placing it. At each step, the model receives the task instruction, current observation, recent interaction context, and admissible commands. These commands include navigation and object manipulation such as taking, placing, opening, closing, toggling, heating, cooling, cleaning, slicing, and inspecting. The environment reports \texttt{won} when all task predicates are satisfied. We assign reward $10$ on this event and zero otherwise, and terminate upon environment completion or after 50 steps. The standard \gbepo{} experiments use only the raw observation interface; they do not expose simulator state or the environment-specific tracked state used in the separate GraphGPO protocol discussed below.

\paragraph{Visual Sokoban.}
Sokoban is a deterministic planning puzzle in which a player pushes boxes onto target squares while avoiding irreversible deadlocks~\citep{junghanns2001sokoban}. Our instances are procedurally generated solvable $6\times6$ boards containing one box and one target. The multimodal policy observes an RGB rendering showing the walls, floor, player, box, and target, and chooses one of four actions: \texttt{Up}, \texttt{Down}, \texttt{Left}, or \texttt{Right}. A move into a wall leaves the state unchanged. Moving toward a box pushes it only when the square beyond it is free; boxes cannot be pulled, so an otherwise legal push can make the puzzle unsolvable. We retain the Gym-Sokoban transition reward for training and define success exactly as placing every box on a target. An episode ends upon success or after 15 steps. Training and evaluation boards are generated from disjoint seed ranges to prevent exact board reuse across the two sets.

\paragraph{Compute and implementation.}
Experiments were run on a single compute node equipped with eight NVIDIA H100 GPUs. Individual Qwen2.5-1.5B, Qwen2.5-VL-3B, and Qwen2.5-7B runs used two, two, and four GPUs, respectively; independent seeds were scheduled concurrently when resources allowed. Training uses PyTorch FSDP, vLLM rollout generation, Ray environment workers, and the verl-agent training framework.

\paragraph{Evaluation metrics.}
For $N$ evaluation episodes, success rate is
\begin{equation}
\operatorname{Success}=\frac{100}{N}\sum_{j=1}^{N}\mathbb I[\text{episode }j\text{ completes its goal}].
\end{equation}
In ALFWorld, completion is the environment's \texttt{won} indicator; category columns use the same metric within each of the six task types, and ``All'' averages over all evaluated tasks. In Sokoban, success requires all boxes to be placed on target squares. WebShop additionally reports
\begin{equation}
\operatorname{TaskScore}=\frac{100}{N}\sum_{j=1}^{N}q_j,
\end{equation}
where $q_j\in[0,1]$ is WebShop's original graded task reward for matching the requested product and attributes. WebShop success is stricter: it is one only when $q_j=1$. For every main result, we first average the three evaluation passes within each training seed, then report the mean and standard deviation across three training seeds. Ablations use one fixed training seed and report variation across three evaluation seeds.

\paragraph{Final hyperparameters.}
Tables~\ref{tab:supp-common-hparams} and~\ref{tab:supp-environment-hparams} list the final method, optimization, and environment settings. The values explored for the main method-specific choices are reported in the ablation tables: GAE $\lambda\in\{0,0.5,0.8,1.0\}$, step weight $w\in\{0.5,1.0,1.5,2.0\}$, Bellman iterations in $\{0,1,3,5,10,20\}$, alternative mean/max/softmax backups, gate thresholds, continuous gate variants, outcome-mixing rules, and gate--GAE recursion variants. We use the final settings consistently across the three environments; the ablations diagnose sensitivity rather than tune each environment separately.

\begin{table*}[!t]
\centering
\small
\setlength{\tabcolsep}{4pt}
\begin{tabular}{llll}
\toprule
Method setting & Value & Optimization setting & Value \\
\midrule
Bellman backup & empirical mean & Optimizer & AdamW \\
Discount $\gamma$ & $0.95$ & Learning rate & $10^{-6}$ \\
GAE $\lambda$ & $0.8$ & AdamW $(\beta_1,\beta_2)$ & $(0.9,0.999)$ \\
Bellman initialization & Monte Carlo & Weight decay & $0.01$ \\
Iteration limit / tolerance & $20$ / $10^{-6}$ & Warmup & none \\
Step normalization & per-task mean/std & PPO epochs per update & $1$ \\
Step weight $w$ & $1.5$ & PPO clip range & $0.2$ \\
Gate / GAE order & hard / post-GAE & Gradient-norm clip & $1.0$ \\
$n_{\min}$ / $b_{\min}$ & $2$ / $2$ & Entropy coefficient & $0.001$ \\
Trusted-state outcome weight $\eta_{\min}$ & $0.5$ & KL loss & low-variance, $0.01$ \\
Invalid graph edges / group skew & retained / disabled & KL in reward & disabled \\
Rollout sampling $(T,p,k)$ & $(1.0,1.0,-1)$ & Evaluation sampling $(T,p,k)$ & $(0.4,1.0,-1)$ \\
\bottomrule
\end{tabular}
\caption{Common final Gated-BEPO and actor-optimization hyperparameters. The same shared optimization settings are used for matched baselines unless their method requires a different advantage estimator.}
\label{tab:supp-common-hparams}
\end{table*}

\begin{table*}[!t]
\centering
\small
\setlength{\tabcolsep}{4pt}
\begin{tabular}{lccccccc}
\toprule
Setting & Train & Eval. & Prompt & Response & Steps & PPO mini/micro & TP \\
\midrule
ALFWorld, Qwen2.5-1.5B & $16$ & $128$ & $2048$ & $512$ & $50$ & $256/32$ & $2$ \\
ALFWorld, Qwen2.5-7B & $16$ & $128$ & $2048$ & $512$ & $50$ & $256/8$ & $4$ \\
WebShop, Qwen2.5-1.5B & $16$ & $128$ & $5120$ & $512$ & $15$ & $64/8$ & $2$ \\
WebShop, Qwen2.5-7B & $16$ & $128$ & $5120$ & $512$ & $15$ & $64/4$ & $4$ \\
Sokoban, Qwen2.5-VL-3B & $32$ & $128$ & $1024$ & $512$ & $15$ & $64/8$ & $2$ \\
\bottomrule
\end{tabular}
\caption{Environment- and backbone-specific settings. Train and Eval.\ are the numbers of task instances per training update and evaluation pass; Prompt and Response are maximum token lengths; Steps is the episode horizon; TP is tensor-parallel degree. All runs use rollout-group size $K=8$ and 150 optimization steps. Qwen3-1.7B transfer experiments use the corresponding environment-specific 1.5B settings and the same Gated-BEPO hyperparameters.}
\label{tab:supp-environment-hparams}
\end{table*}

\section{Supplementary Ablation Tables}

Tables~\ref{tab:gate-ablation}--\ref{tab:maxiter-ablation} collect the supplementary ablation results discussed in the main paper. Detailed definitions and interpretations are provided in the corresponding sections below.


\begin{table}[!t]
\centering
\small
\setlength{\tabcolsep}{1.5pt}
\begin{tabular}{llcc}
\toprule
Factor & Setting & Success & Score \\
\midrule
\multirow{4}{*}{Gate form}
 & Binary & $\mathbf{75.91\pm2.71}$ & $\mathbf{88.32\pm0.89}$ \\
 & Soft $k=3$ & $73.05\pm3.45$ & $87.20\pm1.25$ \\
 & Soft $k=4$ & $72.01\pm2.95$ & $87.76\pm1.35$ \\
 & Soft count+branch & $71.35\pm2.56$ & $86.91\pm0.17$ \\
\midrule
\multirow{4}{*}{GAE $\lambda$}
 & $0.0$ & $64.58\pm4.01$ & $84.98\pm1.37$ \\
 & $0.5$ & $72.14\pm2.75$ & $\mathbf{88.43\pm1.04}$ \\
 & $0.8$ & $\mathbf{75.91\pm2.71}$ & $88.32\pm0.89$ \\
 & $1.0$ & $68.88\pm3.85$ & $85.24\pm1.88$ \\
\midrule
\multirow{4}{*}{Step weight $w$}
 & $0.5$ & $72.01\pm4.40$ & $86.11\pm1.88$ \\
 & $1.0$ & $73.96\pm1.95$ & $87.42\pm0.76$ \\
 & $1.5$ & $\mathbf{75.91\pm2.71}$ & $\mathbf{88.32\pm0.89}$ \\
 & $2.0$ & $71.22\pm1.84$ & $87.88\pm0.65$ \\
\bottomrule
\end{tabular}
\caption{Gate-form, GAE-decay, and step-weight ablations on WebShop (\%).}
\label{tab:diagnostic-ablation}
\end{table}

\begin{table}[!t]
\centering
\small
\setlength{\tabcolsep}{5pt}
\begin{tabular}{lc}
\toprule
Backup & Success (\%) \\
\midrule
Mean & $\mathbf{90.36\pm0.97}$ \\
Max & $65.38\pm1.32$ \\
Softmax & $64.58\pm0.97$ \\
\bottomrule
\end{tabular}
\caption{Bellman-backup ablation on ALFWorld under a fixed training seed, with success reported over three evaluation seeds.}
\label{tab:backup-ablation}
\end{table}

\begin{table}[!t]
\centering
\small
\setlength{\tabcolsep}{3pt}
\begin{tabular}{cccc}
\toprule
Iter. & Max $|\Delta A|$ & Success & Score \\
\midrule
$0$ & -- & $71.09\pm3.24$ & $86.02\pm1.02$ \\
$1$ & $1.57$ & $73.70\pm2.05$ & $88.59\pm0.31$ \\
$3$ & $2.05{\times}10^{-1}$ & $70.96\pm2.08$ & $86.21\pm1.13$ \\
$5$ & $6.59{\times}10^{-2}$ & $72.53\pm3.15$ & $\mathbf{88.69\pm0.38}$ \\
$10$ & $9.86{\times}10^{-3}$ & $75.65\pm2.39$ & $88.47\pm0.59$ \\
$20$ & $3.26{\times}10^{-4}$ & $\mathbf{75.91\pm2.71}$ & $88.32\pm0.89$ \\
\bottomrule
\end{tabular}
\caption{Bellman-iteration ablation on WebShop (\%). Max $|\Delta A|$ measures the maximum absolute change in the step advantage after the final iteration. Iteration 0 uses only the Monte-Carlo initialization.}
\label{tab:maxiter-ablation}
\end{table}

\section{GraphGPO Evaluation Protocols}
We report two ALFWorld comparisons with GraphGPO because they answer different questions. They use the same task family, but they differ in the state representation supplied to graph construction.

\paragraph{Environment-specific state information.}
The original GraphGPO ALFWorld environment constructs graph anchors from the textual observation together with manually tracked agent location, held object and its processing status, remembered object locations, and the sorted admissible-action set. This representation injects environment-specific information that is not present in the raw observation and is intended to reduce collisions between observations that correspond to different underlying states. We use this setting to ask whether GraphGPO is competitive when observation aliasing is actively mitigated. Under this stronger representation, GraphGPO reaches $92.7\%$ success, while \gbepo{} reaches $95.7\%$, average on 3 training seeds.

\paragraph{Matched raw-observation protocol.}
For the controlled comparison in Figure~4 of the main paper, GraphGPO and \gbepo{} use the same standard ALFWorld raw textual observations as graph-state keys. We remove the GraphGPO-specific tracked location, inventory and object-history fields from the graph representation; no method receives an alias-resolving state augmentation. All other training and evaluation choices are aligned with the matched ALFWorld setup. This protocol deliberately retains perceptual aliasing and asks how the credit estimators behave when identical observations may denote different hidden states. GraphGPO nearly fails to learn, whereas \gbepo{} remains effective under this setting.

The environment-specific-state result tests both methods after supplying GraphGPO with a stronger, alias-reducing representation. The raw-observation result controls the observation interface and tests sensitivity to the aliasing that remains in the standard environment. \gbepo{} does not reconstruct the hidden Markov state. Instead, after identical observations are merged for value estimation, its TD residuals remain transition-specific and are propagated along the original trajectories with GAE, allowing different occurrences of the same observation to receive different credit when their successors and returns differ.

\section{Illustration of Observation Aliasing}
Consider an ALFWorld task that requires placing a mug in a cabinet. The agent may arrive at the cabinet either before retrieving the mug or after picking it up. Because the raw textual observation describes the location and visible objects but does not encode whether the agent is holding the mug, these two visits can produce the same graph key $o$. Only the hidden state in which the agent holds the mug can complete the task immediately with $a_{\mathrm{put}}$. Without the mug, the necessary action $a_{\mathrm{retrieve}}$ begins a longer route that retrieves the mug and returns to the cabinet. Suppose the rollouts also contain an incorrect action $a_{\mathrm{wrong}}$ that leads to failure. Merging the two visits gives the following empirical transitions, where $p\leadsto\mathrm{success}$ denotes the remaining retrieve-and-return continuation.

\begin{equation}
\begin{aligned}
o &\xrightarrow{a_{\mathrm{put}}} \mathrm{success},\\
o &\xrightarrow{a_{\mathrm{retrieve}}} p \leadsto \mathrm{success},\\
o &\xrightarrow{a_{\mathrm{wrong}}} \mathrm{failure}.
\end{aligned}
\end{equation}

Let success have score one and failure score zero, and omit common scaling and subsequent standardization because they do not change the signs. With GraphGPO's distance discount $\alpha=0.1$, the direct transition $a_{\mathrm{put}}$ receives score $1$, whereas the longer successful continuation after $a_{\mathrm{retrieve}}$ receives $x=\alpha^d\leq0.1$ for at least one additional graph step $d\geq1$. The incorrect transition receives zero. Centering these scores within the merged node gives advantages $(2-x)/3$, $(2x-1)/3$, and $-(1+x)/3$, respectively. Since $x\leq0.1$, GraphGPO assigns at most $-0.267$ to $a_{\mathrm{retrieve}}$, even though retrieving the mug is necessary when the agent does not hold it. The one-step completion observed in the other hidden state makes the required retrieve-and-return route appear to be an inferior detour.

\gbepo{} instead evaluates each observed transition against the empirical mean Bellman value. Write the three Bellman targets as $1$, $y=\gamma V(p)$, and $0$. With one observation of each transition, the mean fixed point at the merged node is $V(o)=(1+y)/3$, yielding residuals $(2-y)/3$, $(2y-1)/3$, and $-(1+y)/3$. Thus, the retrieval action remains positive whenever its observed continuation has $y>0.5$. For example, with $\gamma=0.95$ and $V(p)=1$, the targets are $1$, $0.95$, and $0$, and the residuals are $0.35$, $0.30$, and $-0.65$. Unlike shortest-path scoring, this comparison does not automatically treat every continuation longer than the aliased shortcut as negative.

The merged node passes the confidence gate because it has distinct observed successors, but this branching is only evidence for a local comparison; it does not prove that the two visits represent the same Markov state. Moreover, the retained outcome branch still anchors each update to whether its complete trajectory succeeds or fails. Although \gbepo{} does not recover the hidden inventory state, mean evaluation and conservative outcome--step mixing still reduce the damage caused when the pre-retrieval and post-retrieval observations are accidentally merged.

\section{Outcome--Step Mixing Variants}
The RQ1 ablation keeps graph construction, Bellman evaluation, advantage normalization, and $w=1.5$ fixed while changing only how outcome and fixed-point advantages are combined. Let $\rho_i\in\{0,1\}$ denote the graph-confidence gate and let $\eta(\rho_i)=\eta_{\min}+(1-\eta_{\min})(1-\rho_i)$. The tested variants are defined below.

\begin{itemize}
    \item \textbf{Outcome only} uses $\widehat A_i=\widehat A_i^{\mathrm{out}}$ and disables the fixed-point branch by setting its weight to zero.
    \item \textbf{Ungated addition} uses $\widehat A_i=\widehat A_i^{\mathrm{out}}+w\widehat A_i^{\mathrm{FP}}$, applying fixed-point credit at every state without a graph-confidence test.
    \item \textbf{Dynamic mixture} is the default \gbepo{} rule $\widehat A_i=\eta(\rho_i)\widehat A_i^{\mathrm{out}}+w\rho_i\widehat A_i^{\mathrm{FP}}$ with $\eta_{\min}=0.5$. Untrusted states use outcome credit only, while trusted states retain half of the outcome term and activate fixed-point credit.
    \item \textbf{Full outcome weight} sets $\eta_{\min}=1$, giving $\widehat A_i=\widehat A_i^{\mathrm{out}}+w\rho_i\widehat A_i^{\mathrm{FP}}$. The gate controls only the fixed-point term.
    \item \textbf{Drop trusted outcome} sets $\eta_{\min}=0$, giving $\widehat A_i=(1-\rho_i)\widehat A_i^{\mathrm{out}}+w\rho_i\widehat A_i^{\mathrm{FP}}$. Trusted states therefore rely entirely on fixed-point credit.
    \item \textbf{Group-skew modulation} multiplies the default outcome weight by $\eta_g=\operatorname{clip}(4p_g(1-p_g),0,1)$, where $p_g$ is the mean binary return across trajectories in rollout group $g$. Its rule is $\widehat A_i=\eta_g\eta(\rho_i)\widehat A_i^{\mathrm{out}}+w\rho_i\widehat A_i^{\mathrm{FP}}$.
\end{itemize}

\begin{table*}[!t]
\centering
\small
\setlength{\tabcolsep}{5pt}
\begin{tabular}{p{0.25\linewidth}p{0.43\linewidth}cc}
\toprule
Variant & Credit assignment rule & Success & Task Score \\
\midrule
Outcome only & $\widehat A^{\mathrm{out}}$ & $62.50\pm1.99$ & $81.38\pm0.48$ \\
Ungated addition & $\widehat A^{\mathrm{out}}+1.5\widehat A^{\mathrm{FP}}$ & $66.41\pm3.38$ & $87.16\pm0.41$ \\
\gbepo{} dynamic mixture & $\eta(\rho)\widehat A^{\mathrm{out}}+1.5\rho\widehat A^{\mathrm{FP}}$ & $\mathbf{75.91\pm2.71}$ & $88.32\pm0.89$ \\
Full outcome weight & $\widehat A^{\mathrm{out}}+1.5\rho\widehat A^{\mathrm{FP}}$ & $73.83\pm2.30$ & $87.26\pm1.01$ \\
Drop trusted outcome & outcome if $\rho=0$; fixed-point only if $\rho=1$ & $75.00\pm1.10$ & $\mathbf{88.53\pm0.55}$ \\
Group-skew modulation & default mixture with group-level success/failure skew & $71.35\pm1.33$ & $87.54\pm0.11$ \\
\bottomrule
\end{tabular}
\caption{Complete outcome--step credit mixing ablation on WebShop (\%). All variants use the same graph construction, fixed-point estimator, normalization, and step weight $w=1.5$. The main paper reports the four component-wise configurations; the final two rows provide the additional outcome-weighting variants.}
\label{tab:supp-mixing-ablation}
\end{table*}

Table~\ref{tab:supp-mixing-ablation} reports the complete results. The default dynamic mixture gives the highest success rate, whereas dropping outcome credit at trusted states gives a slightly higher task score but lower success. Group-skew modulation underperforms the default on both metrics, indicating that rollout-group outcome balance is a weaker reliability signal than local graph evidence.

\section{Gate Threshold Results}
Table~\ref{tab:gate-ablation} reports the non-redundant gate-threshold settings discussed in RQ2 of the main paper. Since $b(s)=|\operatorname{Succ}(s)|\leq n(s)$, the omitted $(1,2)$ setting is equivalent to the default $(2,2)$ setting. It outperforms gates applied to all states, gates based only on repeated visits, and more restrictive thresholds, supporting observed branching as a minimum empirical criterion for activating local credit, rather than as a guarantee that state comparisons are free of aliasing.

\section{Gate Form and Step-Signal Results}
The RQ4 ablation replaces the binary gate with two forms of continuous graph confidence. Let $n(s)$ be the number of observed outgoing transitions from state $s$ and let $b(s)=|\operatorname{Succ}(s)|$. We define the clipped ramp $c(x;k)=\operatorname{clip}((x-1)/(k-1),0,1)$.

The count-ramp gate uses $\rho(s)=c(n(s);k)\mathbb{I}[b(s)\geq2]$. We test $k=3$ and $k=4$, which gradually increase trust with repeated visits but still require an observed branch. The count-and-branch gate uses $\rho(s)=c(n(s);4)c(b(s);3)$, so two successors provide partial confidence and three provide full branch confidence. All remaining settings match the binary-gate configuration.

Table~\ref{tab:diagnostic-ablation} reports the complete gate-form, GAE-decay, and step-weight ablations discussed in the main paper. The binary gate, $\lambda=0.8$, and $w=1.5$ form the final configuration.

The gate is deliberately applied after GAE rather than to each temporal-difference residual. This means it certifies the current use location, not every downstream residual contributing to the propagated target. We therefore compare the default recursion
\begin{equation}
A_t^{\mathrm{post}}=\delta_t+\gamma\lambda A_{t+1}^{\mathrm{post}},
\end{equation}
with residual masking,
\begin{equation}
A_t^{\mathrm{mask}}=\rho(s_t)\delta_t+\gamma\lambda A_{t+1}^{\mathrm{mask}},
\end{equation}
and propagation stopping,
\begin{equation}
A_t^{\mathrm{stop}}=\delta_t+\gamma\lambda\rho(s_{t+1})A_{t+1}^{\mathrm{stop}}.
\end{equation}
The current-state factor $\rho(s_t)$ is still applied during final mixing in all three variants; thus, the alternatives change only the temporal recursion, not the gate definition, Bellman values, outcome credit, or normalization. The main-paper ablation shows that post-GAE gating obtains $75.26\pm3.19$ success and $89.92\pm1.00$ WebShop task score, compared with $72.79\pm2.24$ and $87.49\pm1.21$ for residual masking, and $69.92\pm2.23$ and $87.13\pm0.29$ for propagation stopping. Retaining downstream residuals therefore provides useful delayed credit even when a downstream state does not itself pass the local branching criterion.

\section{Bellman Fixed-Point Results}

\paragraph{Backup operator.}
Table~\ref{tab:backup-ablation} compares mean, max, and softmax Bellman backups on ALFWorld. Mean evaluates the empirical transition distribution, while max and softmax emphasize high-value observed successors. Their substantially lower results indicate that planning-style backups can amplify rare or aliased transitions rather than provide a stable state-centered baseline.

\paragraph{Iteration budget.}

Table~\ref{tab:maxiter-ablation} reports the complete Bellman-iteration ablation corresponding to the convergence analysis in the main paper. Iteration zero uses only the Monte-Carlo initialization, whereas positive budgets apply the mean Bellman backup. The maximum absolute advantage change decreases by more than three orders of magnitude from one to twenty iterations and is already below $10^{-2}$ at ten iterations. Success varies at small intermediate budgets but stabilizes once the value estimate is close to its fixed point: ten and twenty iterations obtain similar results. We therefore use twenty iterations as a conservative default, while the small difference after ten iterations indicates that the graph computation does not require a large iteration budget.

\begin{figure*}[!t]
\centering
\begin{minipage}{0.49\textwidth}
\centering
\includegraphics[width=\linewidth]{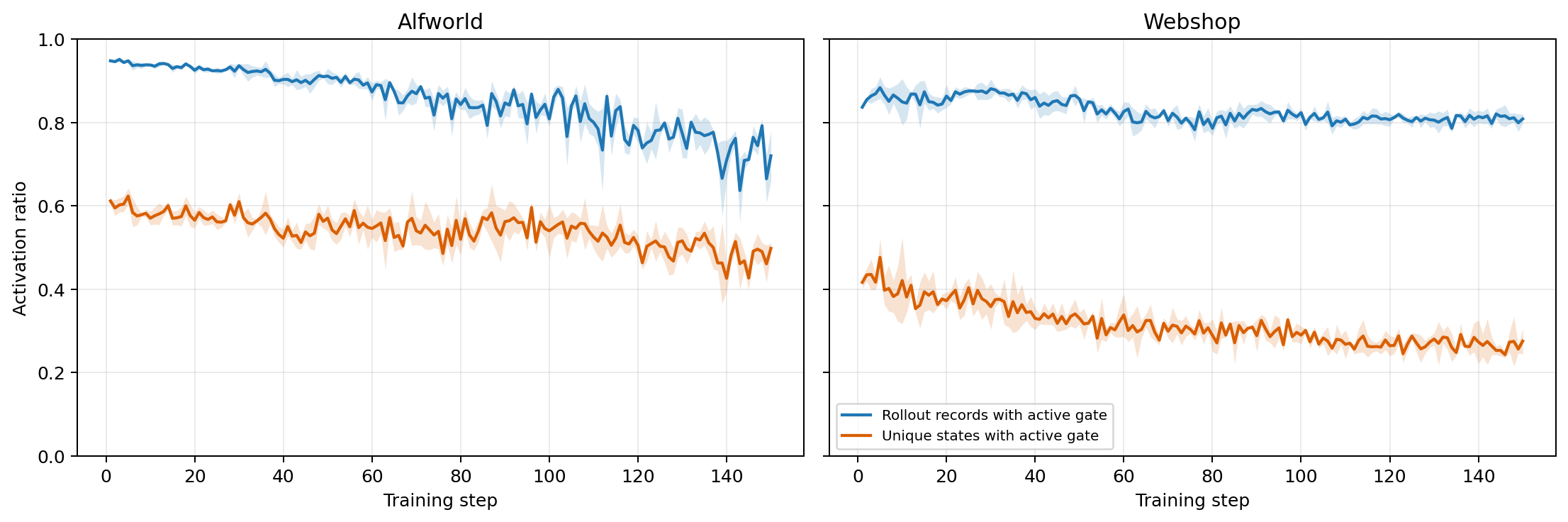}\\[-3pt]
\small (a) Gate activation
\end{minipage}\hfill
\begin{minipage}{0.49\textwidth}
\centering
\includegraphics[width=\linewidth]{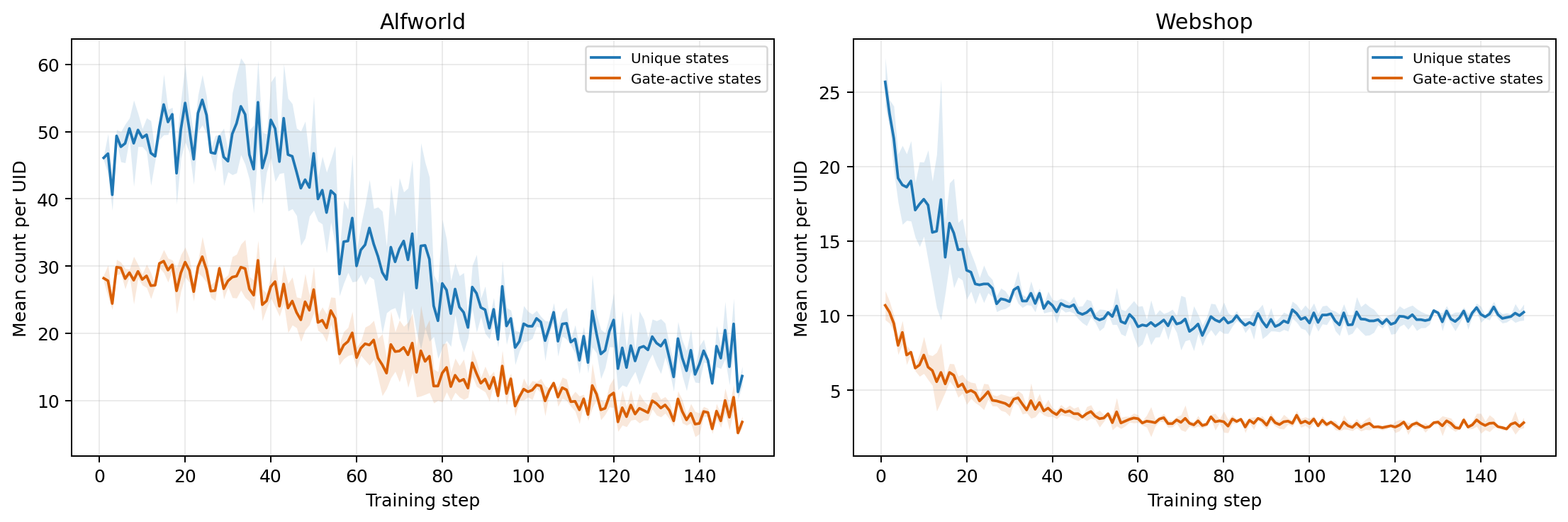}\\[-3pt]
\small (b) Graph-state counts
\end{minipage}\\[2pt]
\begin{minipage}{0.49\textwidth}
\centering
\includegraphics[width=\linewidth]{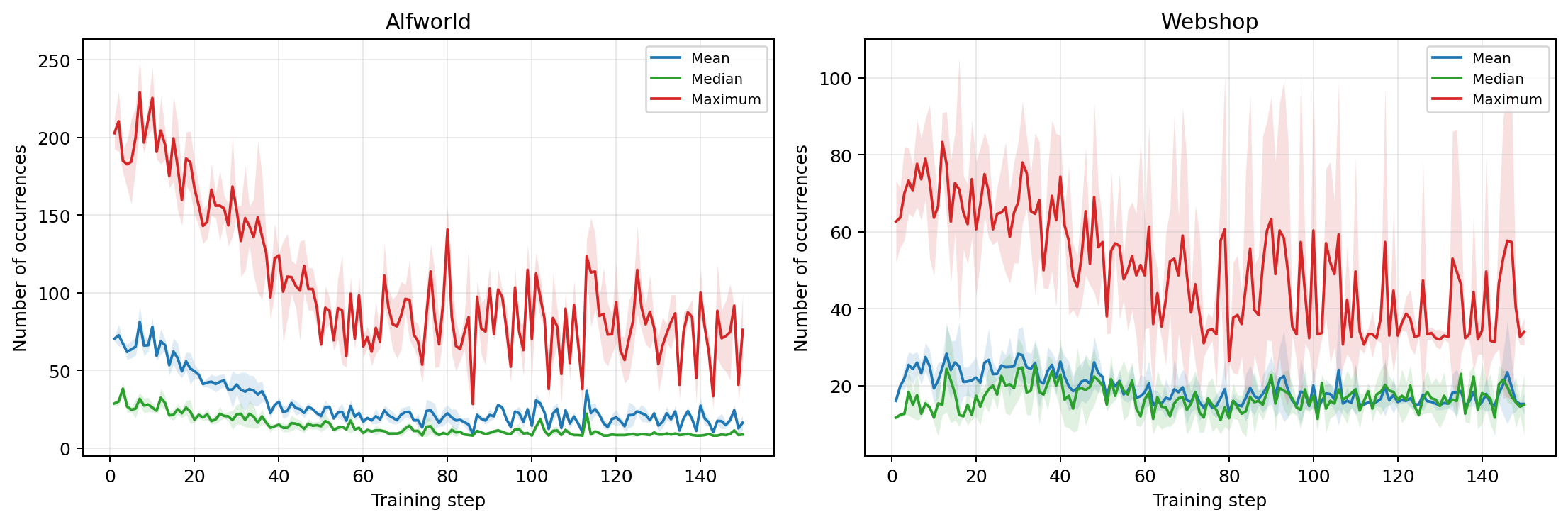}\\[-3pt]
\small (c) State repetition
\end{minipage}\hfill
\begin{minipage}{0.49\textwidth}
\centering
\includegraphics[width=\linewidth]{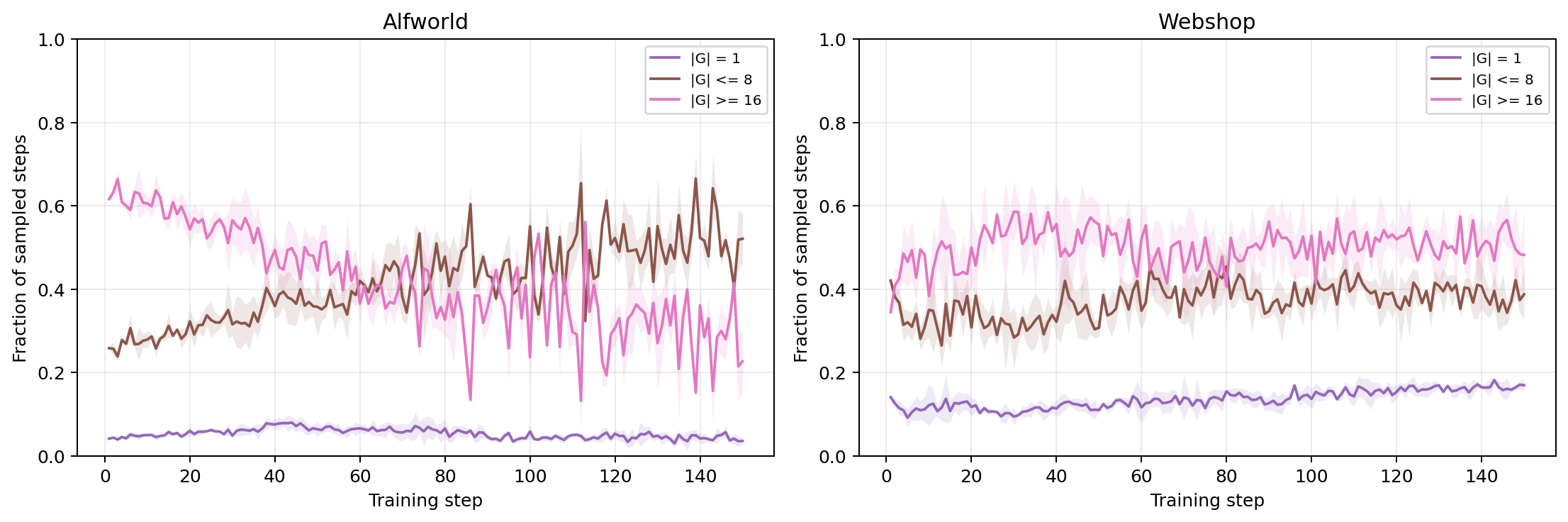}\\[-3pt]
\small (d) Repetition buckets
\end{minipage}
\caption{Shared-state and gate diagnostics over three seeds. Panels report gate coverage, graph size, and within-group state repetition; shaded bands show one standard deviation. Matching and counting are performed independently within each task-level rollout group.}
\label{fig:supp-gate-diagnostics}
\end{figure*}

\section{Shared-State Coverage and Gate Activation}

Figure~\ref{fig:supp-gate-diagnostics} analyzes the empirical support available to the confidence gate. All state matching is scoped to a single task-level rollout group: observations from different task instances are never merged. At the final logged step, the gate is active for $72.0\pm6.0\%$ of ALFWorld rollout records and $80.8\pm1.2\%$ of WebShop records, while the corresponding fractions of unique states are $49.8\pm1.2\%$ and $27.5\pm3.1\%$. Thus, shared states are sufficiently common to affect a substantial fraction of updates, but the gate remains selective over unique graph states. The gap between record- and state-level coverage also shows that frequently revisited states receive proportionally more training weight, making them natural anchors for local transition comparisons.

The graph-size curves provide a complementary view. Both environments visit fewer unique states per task-level group as training proceeds, consistent with the policy concentrating on a smaller set of behaviors. Nevertheless, the record-level activation ratio remains high because the surviving states are repeatedly encountered. This is the regime in which shared-state estimation is most useful: repeated observations provide reusable local evidence, while the distinct-successor requirement prevents mere repetition of a single transition from activating step credit.

\section{Qwen3-1.7B Results}

Table~\ref{tab:qwen3-results} extends the evaluation to Qwen3-1.7B. \gbepo{} gives the strongest result on both environments, exceeding HGPO by $7.29$ success points on ALFWorld and $4.86$ points on WebShop.

The method ordering is identical across ALFWorld and WebShop, so the gain is not tied to one environment interface. On WebShop, \gbepo{} also has the smallest across-seed variation in success among the four methods ($1.52$ points), suggesting that selective step credit remains stable under the Qwen3 policy. These results provide a model-family transfer check rather than a claim that Qwen3 and Qwen2.5 are directly comparable under identical exploration dynamics.

\begin{strip}
\centering
\small
\setlength{\tabcolsep}{6pt}
\begin{tabular}{lccc}
\toprule
Method & ALFWorld Success (\%) & WebShop Success (\%) & WebShop Task Score (\%) \\
\midrule
GRPO & $55.55\pm7.05$ & $49.31\pm11.74$ & $65.99\pm11.22$ \\
GiGPO & $72.30\pm4.90$ & $58.77\pm3.06$ & $80.06\pm2.94$ \\
HGPO & $76.48\pm6.87$ & $62.54\pm3.72$ & $82.83\pm2.28$ \\
\gbepo{} & $\mathbf{83.77\pm4.83}$ & $\mathbf{67.40\pm1.52}$ & $\mathbf{84.65\pm0.94}$ \\
\bottomrule
\end{tabular}
\captionof{table}{Qwen3-1.7B results over three training seeds and three evaluation repeats per seed.}
\label{tab:qwen3-results}
\end{strip}

\section{Additional RL Training Diagnostics}

\begin{figure*}[!t]
\centering
\begin{minipage}{0.32\textwidth}
\centering
\includegraphics[width=\linewidth]{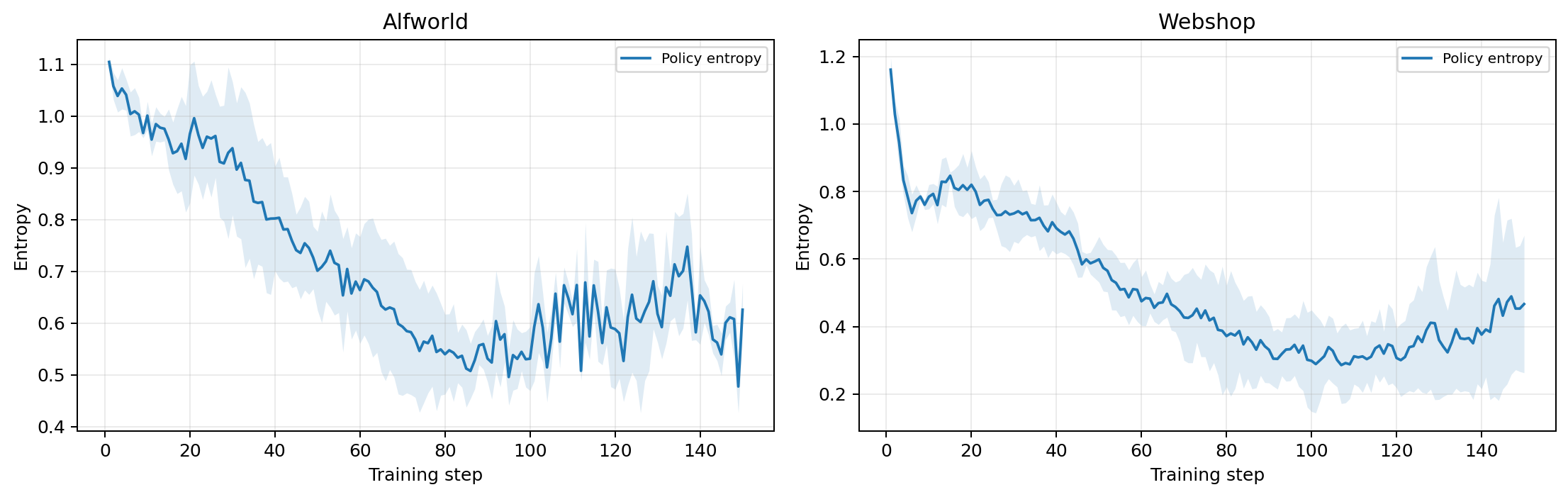}\\[-3pt]
\small (a) Policy entropy
\end{minipage}\hfill
\begin{minipage}{0.32\textwidth}
\centering
\includegraphics[width=\linewidth]{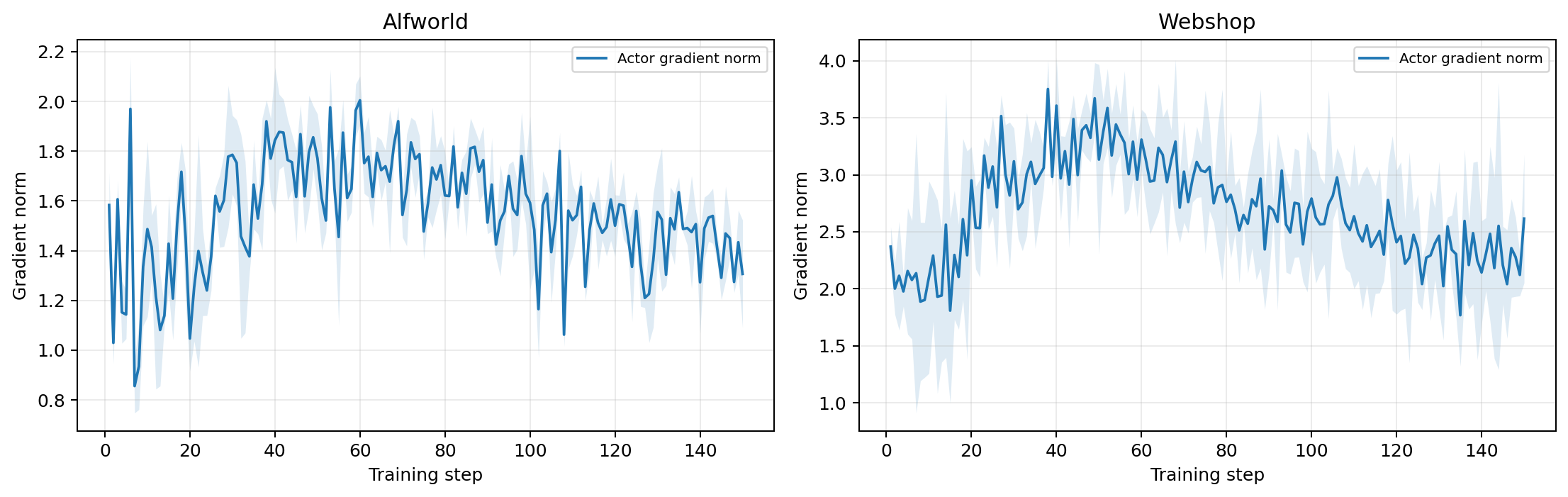}\\[-3pt]
\small (b) Actor gradient norm
\end{minipage}\hfill
\begin{minipage}{0.32\textwidth}
\centering
\includegraphics[width=\linewidth]{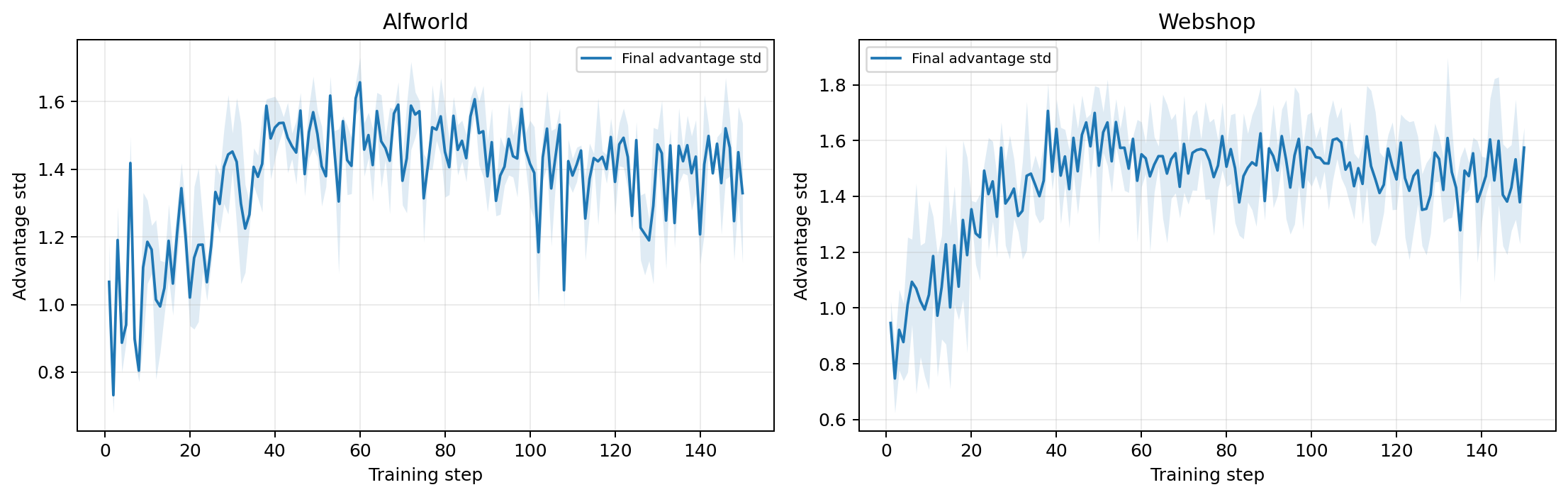}\\[-3pt]
\small (c) Final-advantage standard deviation
\end{minipage}\\[2pt]
\begin{minipage}{0.32\textwidth}
\centering
\includegraphics[width=\linewidth]{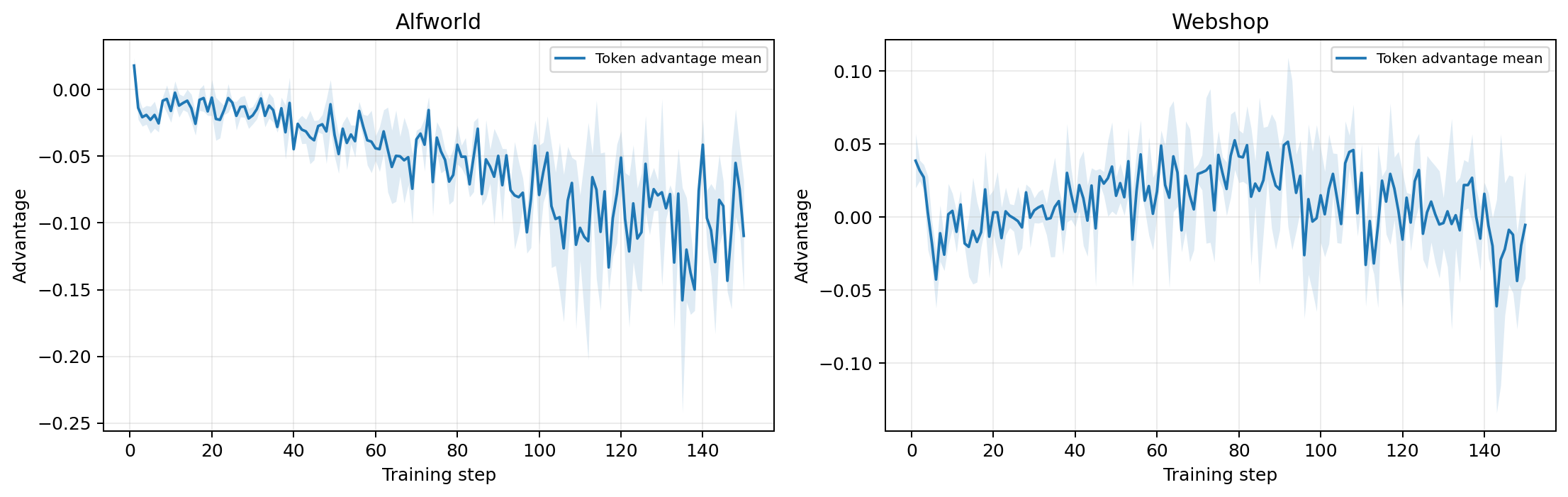}\\[-3pt]
\small (d) Token-advantage mean
\end{minipage}\hfill
\begin{minipage}{0.32\textwidth}
\centering
\includegraphics[width=\linewidth]{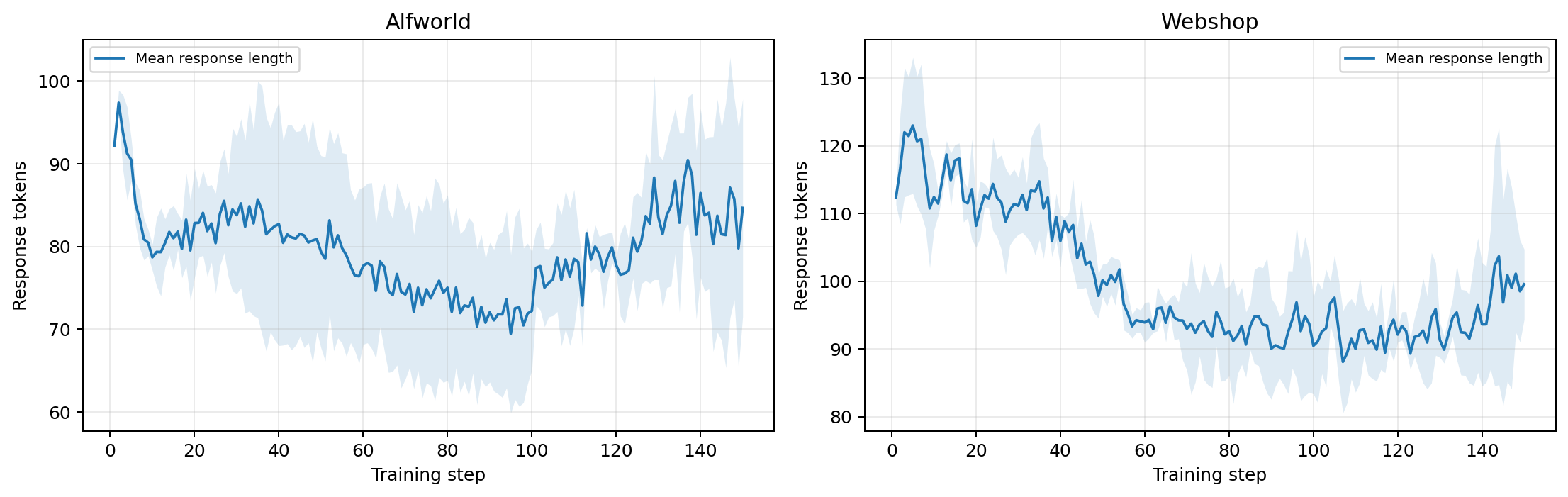}\\[-3pt]
\small (e) Mean response length
\end{minipage}\hfill
\begin{minipage}{0.32\textwidth}
\centering
\includegraphics[width=\linewidth]{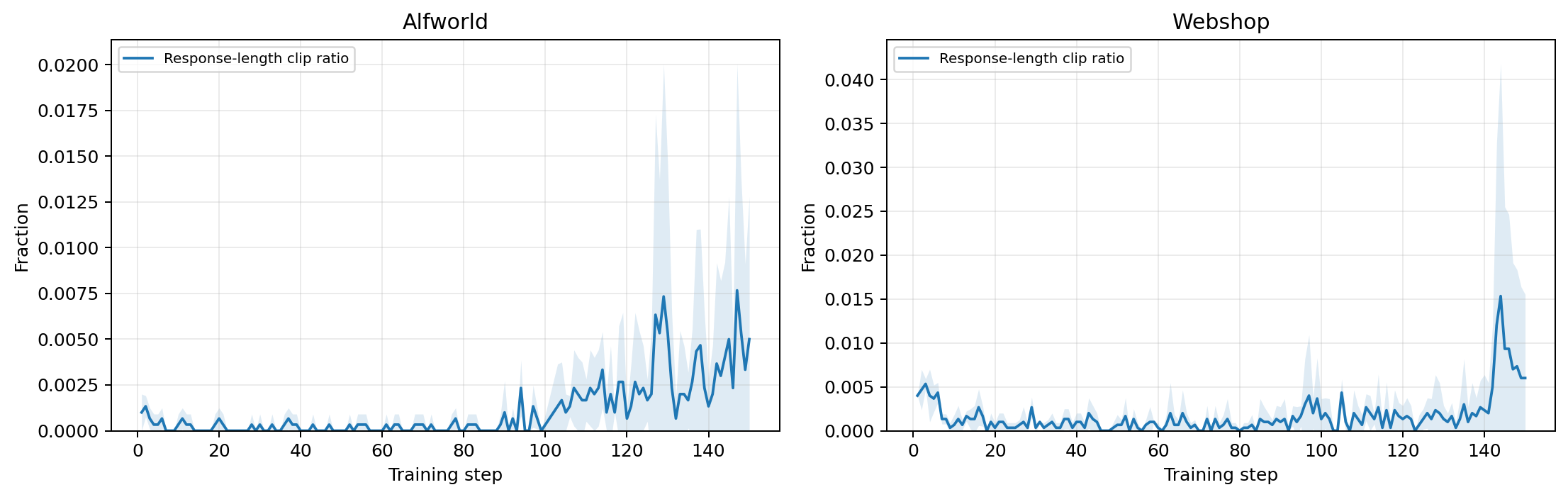}\\[-3pt]
\small (f) Response-length clipping
\end{minipage}
\caption{RL training diagnostics for the main Gated-BEPO runs. Each panel contains ALFWorld and WebShop traces and reports the mean and one-standard-deviation band over three seeds.}
\label{fig:supp-rl-signals}
\end{figure*}

Figure~\ref{fig:supp-rl-signals} reports common optimization, advantage, and response-length statistics for the main ALFWorld and WebShop runs. Policy entropy decreases gradually rather than collapsing at the beginning of training, while the gradient norm remains bounded after a mid-training rise. The final record-level advantage standard deviation stabilizes after the initial phase, indicating that the mixed signal does not continually grow in scale.

The token-expanded advantage mean stays small relative to its standard deviation but need not be exactly zero, because record-level normalization precedes broadcasting and variable response lengths reweight records. Response lengths evolve smoothly, with rare clipping until a modest late increase. Together, the traces show no evidence of exploding updates, degenerate advantage scaling, or pervasive truncation.

\section{Additional Learning Curves}

\begin{figure}[!t]
\centering
\includegraphics[width=\linewidth]{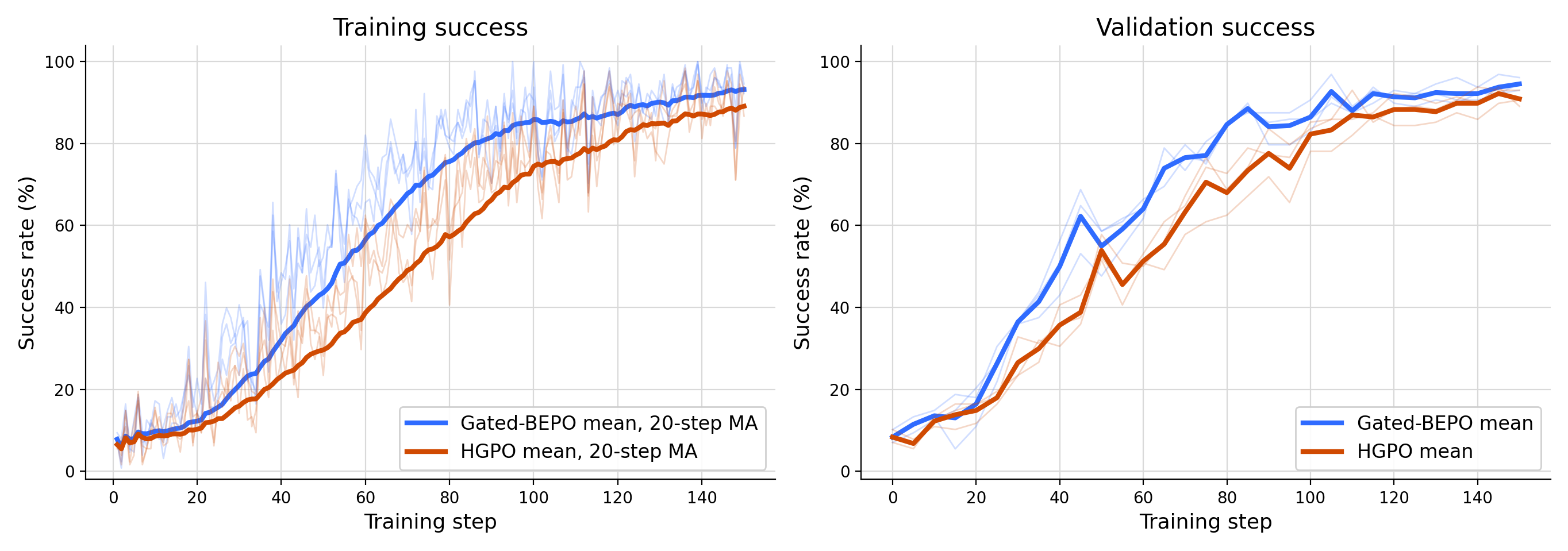}
\caption{ALFWorld training and validation success-rate curves for \gbepo{} and HGPO with Qwen2.5-1.5B under the locally matched raw-observation protocol. Lines show means across three training seeds, and shaded bands show one sample standard deviation across seeds.}
\label{fig:supp-alfworld-1p5b-validation-curves}
\end{figure}

Figure~\ref{fig:supp-alfworld-1p5b-validation-curves} compares the ALFWorld training and validation success of \gbepo{} and HGPO with Qwen2.5-1.5B. \gbepo{} improves more rapidly than HGPO and retains the highest validation success near the end of training.

\begin{figure}[!t]
\centering
\includegraphics[width=\linewidth]{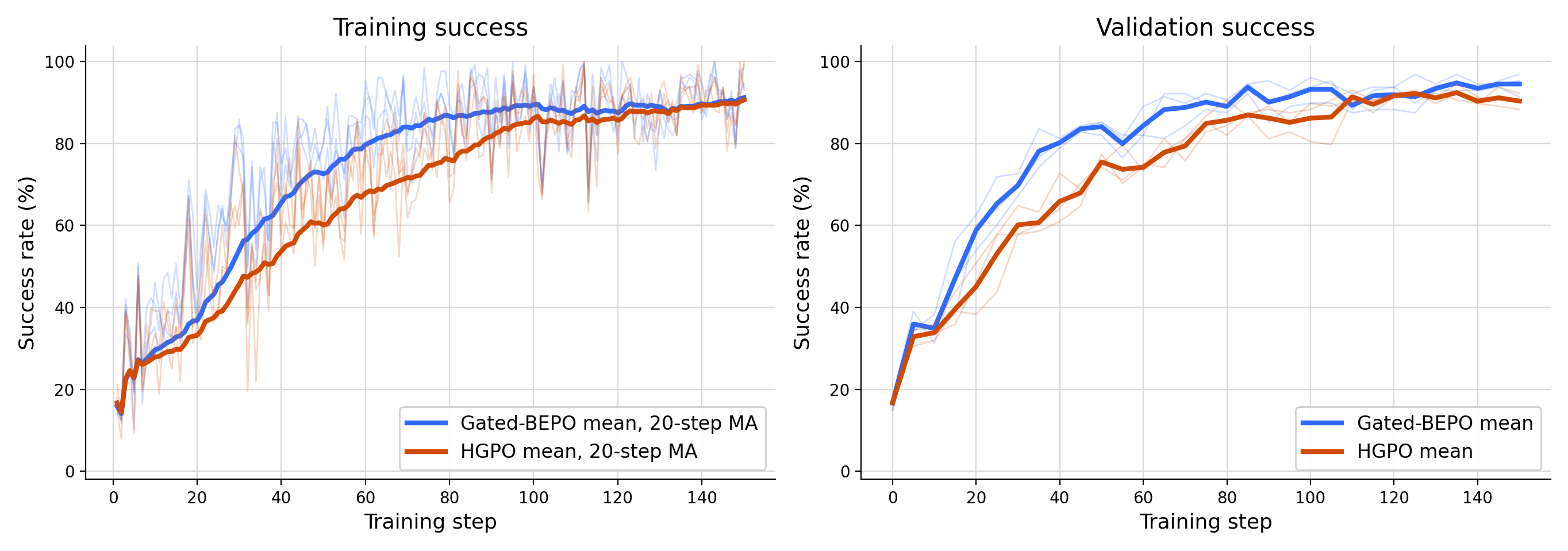}
\caption{ALFWorld training and validation success-rate curves for \gbepo{} and HGPO with Qwen2.5-7B under the locally matched raw-observation protocol.}
\label{fig:supp-alfworld-7b-curves}
\end{figure}

Figure~\ref{fig:supp-alfworld-7b-curves} reports the ALFWorld learning curves for Qwen2.5-7B. The two methods eventually reach similar training success, while \gbepo{} retains higher validation success near the end of training.

\section{Algorithm Details}

Algorithm~\ref{alg:gated-bepo} summarizes the advantage computation used by \gbepo{}. All graph construction, value estimation, and normalization are performed independently within each task-level rollout group.

\begin{algorithm}[H]
\caption{Gated-BEPO Advantage Computation}
\label{alg:gated-bepo}
\begin{algorithmic}[1]
\REQUIRE Rollout groups $\mathcal U$ containing trajectories $\tau$ and returns $R(\tau)$
\REQUIRE $\gamma$, $\lambda$, $w$, $\eta_{\min}$, $n_{\min}$, $b_{\min}$
\ENSURE Token-level advantages for PPO
\FOR{each rollout group $u\in\mathcal U$}
  \STATE Let $\mathcal I_u$ contain all transition records in the trajectories of $u$.
  \STATE Compute group-relative $\widehat A_i^{\mathrm{out}}$ from $R(\tau_i)$ over $i\in\mathcal I_u$.
  \STATE Map observations to states and terminal outcomes to zero-valued absorbing states.
  \STATE Build $E_u(s)=\{(r_i,s'_i):i\in\mathcal I_u,\ s_i=s\}$ with multiplicities.
  \STATE Initialize $V$ with Monte-Carlo returns.
  \STATE Iterate $V(s)\gets |E_u(s)|^{-1}\sum_{(r,s')\in E_u(s)}[r+\gamma V(s')]$ until convergence or the iteration limit.
  \STATE Set $\delta_i\gets r_i+\gamma V(s'_i)-V(s_i)$ for every $i\in\mathcal I_u$.
  \FOR{each trajectory $\tau\in u$}
    \STATE Compute $\widehat A_i^{\mathrm{FP}}\gets\delta_i+\gamma\lambda\widehat A_{\operatorname{next}(i)}^{\mathrm{FP}}$ backward along $\tau$.
  \ENDFOR
  \STATE Standardize $\{\widehat A_i^{\mathrm{FP}}:i\in\mathcal I_u\}$ over the full group.
  \STATE For each state, compute $n(s)$, $\Succ(s)$, and $\rho(s)=\mathbb I[n(s)\geq n_{\min}\land|\Succ(s)|\geq b_{\min}]$.
  \STATE Set $\eta_i\gets\eta_{\min}+(1-\eta_{\min})(1-\rho(s_i))$ for every $i\in\mathcal I_u$.
  \STATE Set $\widehat A_i\gets\eta_i\widehat A_i^{\mathrm{out}}+w\rho(s_i)\widehat A_i^{\mathrm{FP}}$ and broadcast it to valid response tokens.
\ENDFOR
\RETURN Token-level advantages
\end{algorithmic}
\end{algorithm}

\section{Runtime Analysis}

\begin{figure}[H]
\centering
\includegraphics[width=\columnwidth]{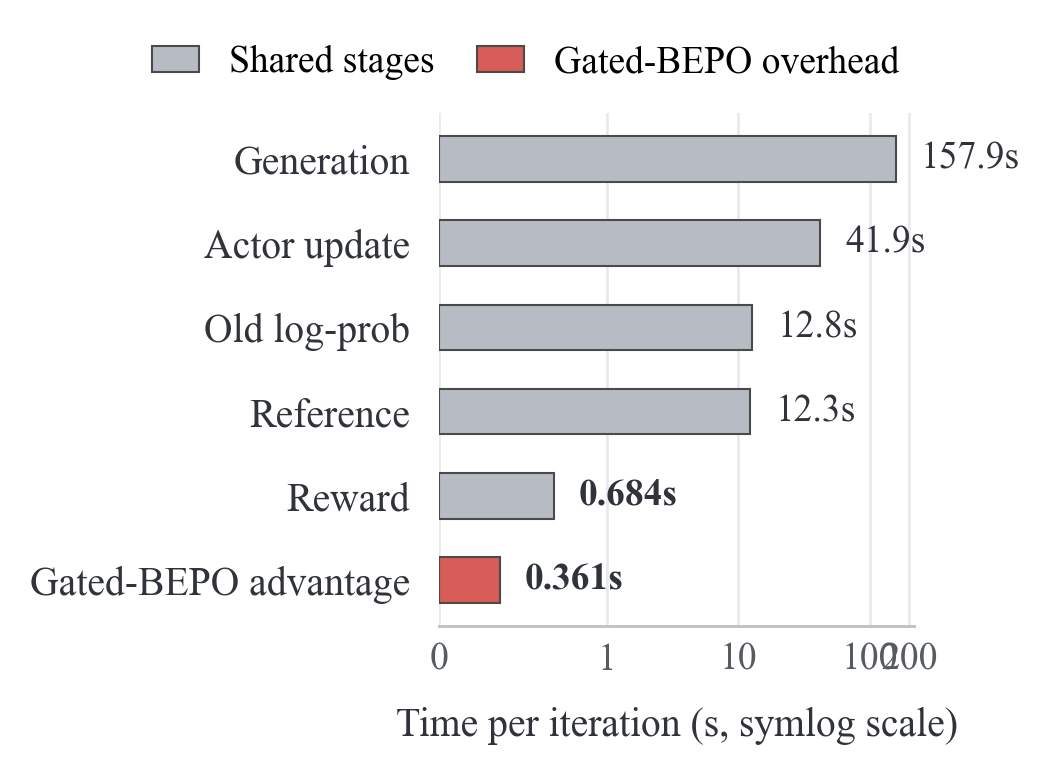}
\caption{ALFWorld per-step runtime on a symmetric-log scale. Gray bars show shared stages and the red bar shows Gated-BEPO advantage computation. Times are measured from the three-seed training logs.}
\label{fig:supp-runtime}
\end{figure}

Figure~\ref{fig:supp-runtime} reports the ALFWorld wall-clock breakdown from the same three-seed main runs. Complete advantage computation takes $0.361$ seconds per update, approximately $0.16\%$ of total step time, while generation and actor optimization dominate the runtime.

The measured advantage stage includes empirical-graph construction, fixed-point evaluation, credit mixing, normalization, and diagnostics. Its cost is negligible relative to rollout and policy optimization, although absolute timings remain hardware- and implementation-dependent.

\end{document}